\documentclass[twocolumn,10pt]{IEEEtran}
\usepackage[mathscr]{eucal}
\usepackage{ifpdf}
\usepackage{cite}
\usepackage{graphicx}
\usepackage[cmex10]{amsmath}
\usepackage{amssymb}
\usepackage{amsmath}
\usepackage{placeins}

\DeclareMathOperator*{\argmin}{arg\,min}
\usepackage{algorithmic}
\usepackage{units}
\usepackage{setspace}
\usepackage{algorithm}
\usepackage{array}
\usepackage{booktabs}
\usepackage{amsthm}
\usepackage{multirow}
\usepackage{enumerate}
\usepackage{color}
\usepackage{mathtools}
\usepackage{caption}
\usepackage[labelformat=simple]{subcaption}
\usepackage{tabularx}
\usepackage{tcolorbox}
\usepackage{supertabular}
\usepackage{tikz}
\usepackage{bm}
\usepackage{xcolor,colortbl}
\usepackage{titlesec}
\usepackage{enumitem}
\usepackage{hyperref}
\usepackage{xfrac}
\usepackage{stfloats}
\usetikzlibrary{arrows.meta, positioning,patterns}

\usepackage{pifont}

\newcounter{assume}

\newcommand{\resl}[1]{}
\newcommand{\tosl}[1]{}
\newcommand{\cosl}[1]{}

\definecolor{mygreen}{RGB}{143, 188, 187}
\definecolor{green1}{RGB}{147, 196, 125}
\definecolor{red1}{RGB}{213,88,88}

\begin{document}
\title{Domain Generalization in Machine
Learning Models for Wireless Communications: Concepts, State-of-the-Art, and Open Issues}
\author{
\IEEEauthorblockN{Mohamed Akrout, Amal Feriani, Faouzi Bellili, \textit{Member, IEEE},\\ Amine Mezghani, \textit{Member, IEEE}, and Ekram Hossain, \textit{Fellow, IEEE}}
}
\maketitle
%
\begin{abstract}
Data-driven machine learning (ML) is promoted as one potential technology to be used in next-generations wireless systems. This led to a large body of research work that applies ML techniques to solve problems in different layers of the wireless transmission link. However, most of these applications rely on supervised learning which assumes that the source (training) and target (test) data are independent and identically distributed (i.i.d). This assumption is often violated in the real world due to domain or distribution shifts between the source and the target data. Thus, it is important to ensure that these algorithms generalize to out-of-distribution (OOD) data. In this context, domain generalization (DG) tackles the OOD-related issues by learning models on different and distinct source domains/datasets with generalization capabilities to unseen new domains without additional finetuning. Motivated by the importance of DG requirements for wireless applications, we present a comprehensive overview of the recent developments in DG and the different sources of domain shift. We also summarize the existing DG methods and review their applications in selected wireless communication problems, and conclude with insights and open questions.

\let\thefootnote\relax\footnotetext{The authors are with the Department of Electrical 
and Computer Engineering, University of Manitoba, Winnipeg, MB, Canada (e-mails: 
\{akroutm,feriania\}@myumanitoba.ca, \{faouzi.bellili, amine.mezghani, ekram.hossain\}@umanitoba.ca). The work was supported by a Discovery Grant from the Natural Sciences and Engineering Research Council of Canada (NSERC).}
\end{abstract}


\section{Introduction}
\label{Sec:Intro}

\subsection{Motivation}

The envisioned design, standardization\footnote{See 3GPP Release 18 \cite[Section 9.2]{3GPPRelease18} for some potential use cases of ML in wireless.}, and deployment of ML in wireless networks require the establishment of evaluation guidelines to properly assess the true potential of data-driven methods. Nevertheless, almost all the openly published ML techniques for wireless systems have several limitations such as $i)$ difficulty to generalize under a \emph{distribution shift}, $ii)$ inability to continuously learn from different scenarios, and $iii)$ inability to \emph{quickly} adapt to unseen scenarios, to name a few. Their showcased performance seems over-fitted to a specific set of simulation settings or fixed datasets, thereby limiting their attractiveness to compete with classical methods at the moment. As one example, the linear minimum mean-square error (LMMSE) estimator of an arbitrary channel model is considered by the industry as one of the most robust estimators in practice. While it is always possible to beat the LMMSE estimator with deep neural network (DNNs) approximators \cite{DBLP:journals/wc/BelgiovineSBRC21}, this fact holds only for an \textit{a priori} known model that is used to generate training and test datasets on which DNNs are trained and also evaluated. When the distributions of training and test datasets are different (e.g., Ricean vs. Rayleigh, or sparse vs. rich-scattering channels), the performance of DNNs deteriorates appreciably due to domain distribution gaps. Furthermore, the lack of real-world wireless communication datasets aggravates the uncertainty toward the practical deployment of ML-based methods. This calls for the development of new ML training algorithms and the establishment of rigorous evaluation protocols to assess their OOD generalization.

In this work, we focus on generalization under domain shift. This includes any change in the distribution between the training (i.e., source) data and the target (i.e., test data). The most studied type of distribution shift is \emph{covariate} shift when the distribution of the model inputs (or features) changes between the source and the target domains \cite{DBLP:conf/ijcai/LiuHJHXOLFW21}. It was shown that the performance of DNNs degrades drastically due to small variations or perturbations in the training datasets \cite{DBLP:journals/corr/abs-1903-12261}. Thus, the acclaimed success of deep learning (DL) is mostly driven by the power of supervised learning. One straightforward idea to overcome domain shift is to adapt the model to the new domain via additional finetuning using techniques such as transfer learning \cite{DBLP:journals/pieee/ZhuangQDXZZXH21} and domain adaptation \cite{DBLP:journals/ijon/WangD18}. However, this is not always feasible in practice because $i)$ target \emph{labeled} data may not be available for finetuning and $ii)$ the finetuning or adaptation may take a long time in contrast to the ``real time'' requirement in most wireless applications. This motivates the DG problem \cite{DBLP:journals/corr/abs-2103-02503} to handle domain shift \textit{without} requiring target domain data. 

DG has been extensively studied in the last decade in the ML community which led to a broad spectrum of methodologies and learning techniques. Moreover, DG was examined in different applications, namely, computer vision \cite{DBLP:conf/iccv/LiYSH17,DBLP:conf/icml/LiYZH19}, natural language processing \cite{DBLP:conf/icml/MillerKRS20,DBLP:conf/acl/JoshiH22}, and medical imaging \cite{DBLP:journals/cbm/LiLMLQDHLY22}, etc. Here, we emphasize the importance of the DG problem in wireless applications to advance the current state-of-the-art research, and raise attention to the problem of domain shift which can seriously impede the success of ML techniques in wireless networks. Specifically, we highlight the importance of leveraging wireless communication domain knowledge to tailor or design more generalizable ML algorithms. 

This work provides a timely and comprehensive overview of the DG research landscape and insights into promising future research directions. The scope of this paper is limited to the DG problem as defined above. At the time of the writing, we have identified several DG variants proposed in the literature that we will briefly discuss but we focus on the standard definition of the DG problem. Other related fields such as domain adaptation, transfer learning, zero-shot learning, multi-task learning, and test time training are beyond the scope of this work. However, we will explain the difference between these fields and DG. In addition, we do distinguish between the terms ``generalization'' and ``robustness'', unlike most wireless communication papers which use them interchangeably. Here, generalization which is also known as \emph{model robustness} denotes the ability of DNNs to generalize to unseen scenarios under distribution shifts. Robustness, however, refers to the stability of DNNs' performance under noise and adversarial examples, i.e., \emph{adversarial robustness} \cite{DBLP:journals/corr/abs-2007-00753}.

\subsection{Contributions and Organization of the Paper}
The main contributions of this paper are summarized as
follows:
\begin{itemize}
    \item We define the DG problem and present four types of distribution shifts. We then contrast DG to existing research fields such as domain adaptation, transfer learning, continual learning, etc.
    \item We summarize different ML methodologies for DG which focus on the following three DNN training steps: (i) data manipulation to cover richer domains pertaining to a given dataset, (ii) representation learning to acquire domain-invariant features enabling generalization, and (iii) learning frameworks which go beyond the standard gradient-based DNN optimization.
    \item We also review the literature on previous attempts for applying ML techniques for DG in several wireless communications problems such as channel decoding, beamforming, multiple-input multiple-output (MIMO) channel estimation, and reconfigurable intelligent  surface (RIS)-aided communications. To the best of our knowledge, this is the first initiative to reconsider the existing applications of ML techniques in wireless research from the DG perspective.
    \item We present the main challenges facing the application of data-driven machine learning techniques in wireless communication under DG requirements and discuss their potential for improving the network performance.
\end{itemize}

\begin{figure}[!ht]
     \centering
         \centering
         \includegraphics[scale=0.47]{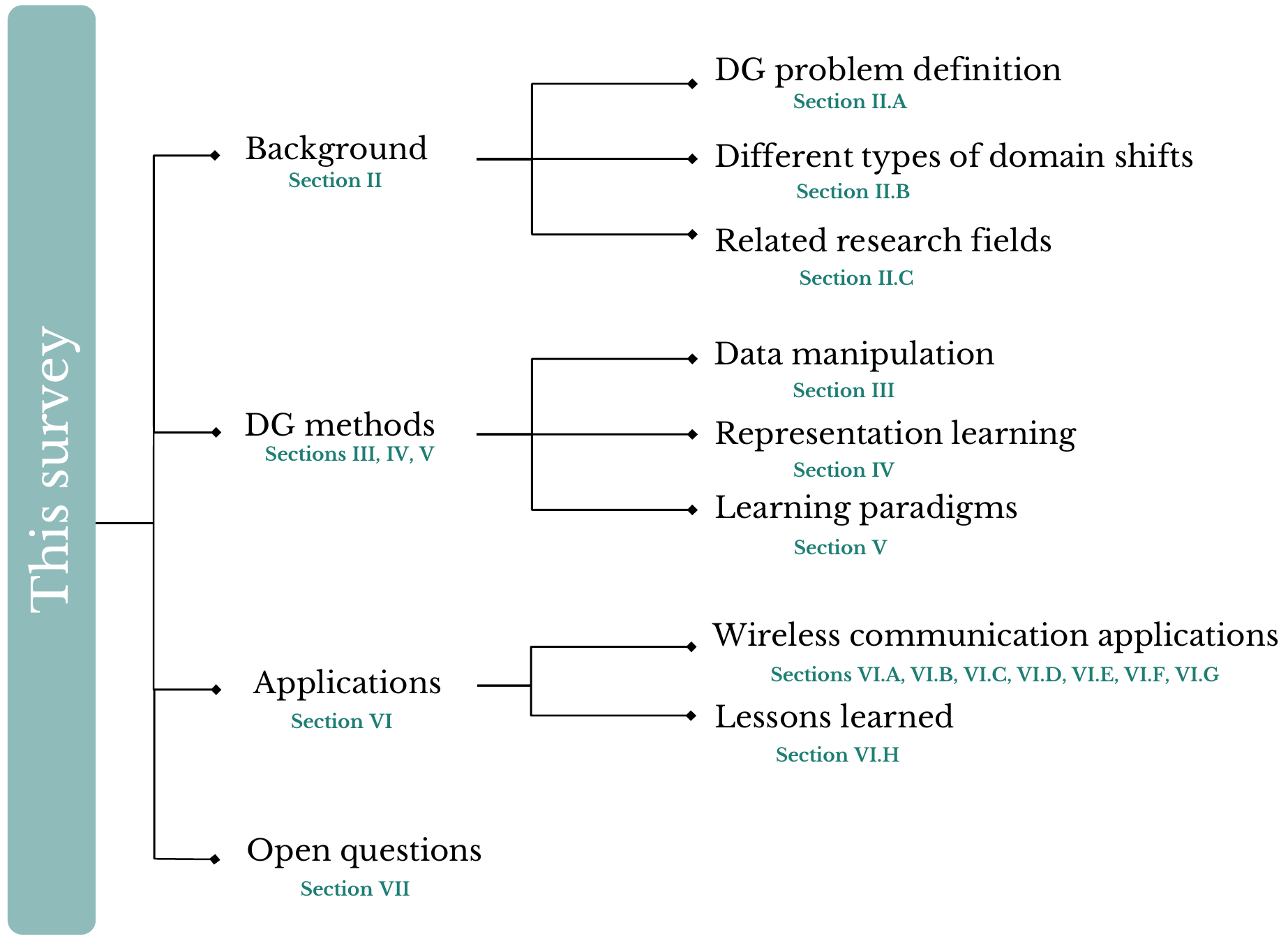}
         \caption{Scope of this work.}
         \label{fig:scope-paper}
\end{figure}

The rest of the paper is organized as illustrated in Fig. \ref{fig:scope-paper}. In Section \ref{sec:background}, we introduce the DG problem formulation and show its key differences with related research fields. State-of-the-art algorithms for DG belonging to data manipulation, representation learning, and learning paradigms are reviewed in Sections \ref{sec:data-manipulation}, \ref{sec:representation-learning}, and \ref{sec:learning-paradigms}, respectively. Section \ref{sec:future-direction-applications} showcases the recent advances of the reviewed DG algorithms in several wireless communication problems, followed by a summary of the learned lessons from their applications. Finally, we outline in Section \ref{sec:future-direction} potential research directions, from which we draw out our concluding remarks in Section \ref{sec:conclusion}.

\section{Background}\label{sec:background}
In this section, we will define the scope of the tutorial and highlight the resemblance between domain generalization and other related research fields. We start by introducing the following definitions.
\subsection{DG Problem Formulation}
\noindent \textbf{Definition 1. (domain)}\label{def:domain}
Let $\mathcal{X}$ and $\mathcal{Y}$ be the input space and the output space of a dataset $\mathcal{D}=\{(X_i,Y_i)\big|P_{\textsf{X},\textsf{Y}}(X_i,Y_i)\}_{i=1}^{K}$, where $K$ is the size of the dataset, $X_i\in\mathcal{X}$ and $Y_i\in\mathcal{Y}$ are the $i$-th input and label samples, respectively. When $X_i$ and $Y_i$ are seen as realizations of their respective random variables $\mathsf{X}$ and $\mathsf{Y}$, it is possible to define a domain as their joint distribution $P_{\textsf{X},\textsf{Y}}(X,Y)$. Moreover, $P_\mathsf{X}(X)$ and $P_\mathsf{Y}(Y)$ refer to the marginal distribution of $\mathsf{X}$ and $\mathsf{Y}$, respectively. Throughout the paper, we will drop distribution arguments to lighten the notation.

Machine learning algorithms use one or multiple datasets, and as such make use of one or multiple data domains according to Definition \hyperref[def:domain]{1}. Indeed, training and evaluating ML techniques require \textit{at least} two domains:
\begin{itemize}[leftmargin=*]
    \item a source (e.g., training) domain $P_{\textsf{X},\textsf{Y}}^{s}$ encoding both the source input marginal $P_{\textsf{X}}^s$ and the source label marginal $P_{\textsf{Y}}^s$;
    \item a target (e.g., test) domain $P_{\textsf{X},\textsf{Y}}^t$ encoding both the target input marginal $P_{\textsf{X}}^t$ and the target label marginal $P_{\textsf{Y}}^t$.
\end{itemize}

Generalization is an active ML research area where the ultimate goal is to learn models that perform well on unseen data domains. This tutorial focuses on out-of-distribution generalization or domain generalization (DG). In the next section, we define the DG problem and explain the difference between this subfield and other generalization problems in ML.\vspace{0.5cm}

\noindent\textbf{Definition 2. (Domain Generalization)}\label{def:DG}  The traditional setting of DG consists of $M$ \emph{distinct} source datasets, i.e., $\mathcal{D}_{\textrm{train}} = \{\mathcal{D}^s\}_{s=1}^M$ with $\mathcal{D}^s = \{(X^s_i,Y^s_i, d^s_i)\big|P_{\textsf{X}^s,\textsf{Y}^s}(X_i^s,Y_i^s)\}_{i}$. Here, the $i$th data-target sample pair $(X_i^s, Y_i^s)$ is sampled from the domain  $P^s_{\textrm{XY}}$ pertaining to the dataset $\mathcal{D}^s$, i.e., $ (X_i^s, Y_i^s) \sim P^s_{\textrm{XY}}$, and $d_i^s$ is a label that is used to distinguish the key characteristics of the domain to which the data-target samples belong, e.g., radar, mmWave transmission, etc. DG also considers unseen target (i.e., test) datasets $\mathcal{D}_{\textrm{test}} = \{\mathcal{D}^t\}_{t=1}^T$ which are different from the source datasets (i.e., $\mathcal{D}^t= \{(X^t_j, Y^t_j)\big|P_{\textsf{X}^t,\textsf{Y}^t}(X_i^t,Y_i^t)\}_{j}  \neq \mathcal{D}^s, ~\forall s$ for $1 \leq s \leq M$). The goal of DG is to train on the source domains a model $f$ that generalizes to the target domains \textit{without any access to the target data during training}. The generalization is often measured via a loss function $\mathcal{L}(\cdot,\cdot)$ on the test domains, i.e., $\mathbb{E}_{(\mathsf{X}_j^t, \mathsf{Y}_j^t) \in \mathcal{D}^t} \left[\mathcal{L}\big(f(X_j^t), Y_j^t\big)\right]$.
\noindent Different variations of the vanilla DG described above have been studied in the literature:
\begin{itemize}
    \item \textbf{Single-source DG} assumes that the training data is \emph{homogeneous} and belongs to a single domain (i.e., $M=1$);
    \item \textbf{Homogeneous DG} requires the source and target domains to share the same label space, i.e., $\mathcal{Y}^s = \mathcal{Y}^t$;
    \item \textbf{Heterogeneous DG} assumes different label spaces for the source and target domains, i.e., $\mathcal{Y}^s \neq \mathcal{Y}^t$;
    \item \textbf{Compound DG}: The vanilla DG setting assumes that source domain labels $d^s$ are known prior to learning. In contrast, compound DG does not require domain annotations and assumes that the source data is \emph{heterogeneous} and consists of mixed domains. In other words, the training is not divided into distinct domains before learning. Thus, in addition to generalizing to new unseen domains, compound DG methods need to infer/learn domain information from mixed heterogeneous datasets. For this reason, compound DG is more challenging than vanilla DG. 
\end{itemize}

\begin{figure*}[th!]
     \centering
     \begin{subfigure}[b]{0.49\textwidth}
         \centering
         \includegraphics[scale=0.43]{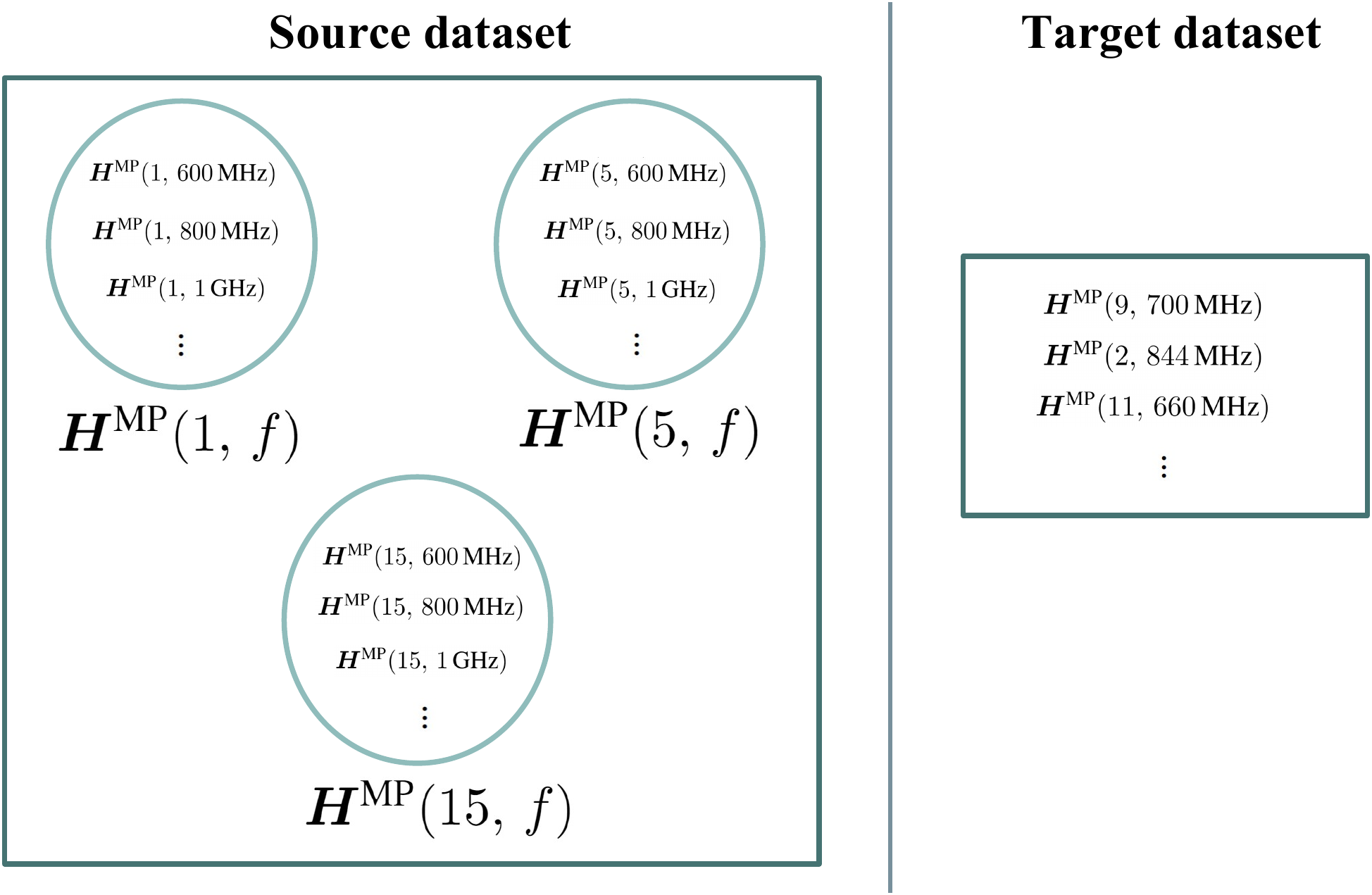}
         \caption{Vanilla DG for the multi-path channel domain $\bm{H}^{\textrm{MP}}(L,\,f)$}
         \label{fig:vanilla-dg}
     \end{subfigure}
     \begin{subfigure}[b]{0.49\textwidth}
         \centering
         \includegraphics[scale=0.43]{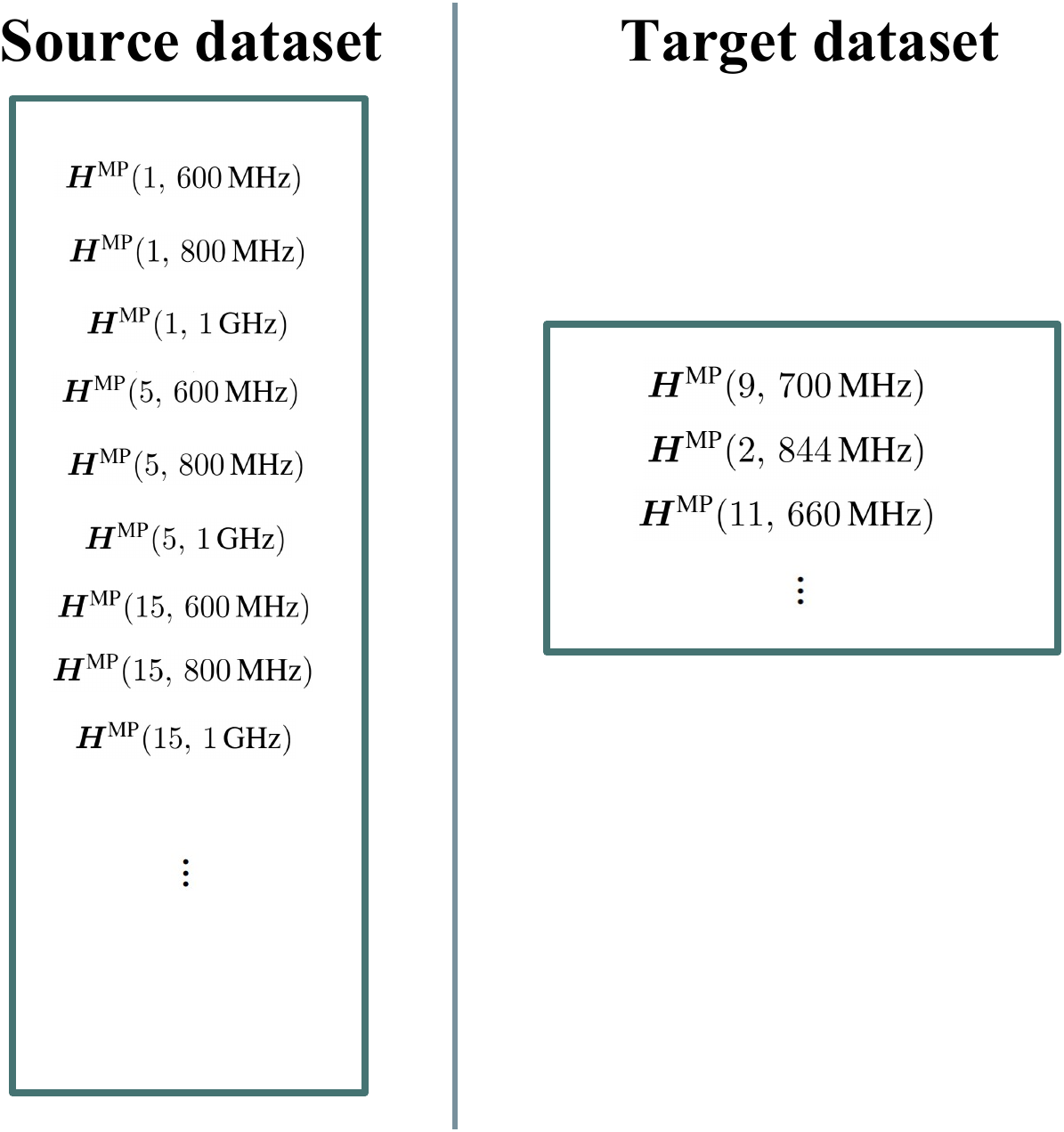}
         \caption{Compound DG for the multi-path channel domain $\bm{H}^{\textrm{MP}}(L,\,f)$}
         \label{fig:compound-dg}
     \end{subfigure}
    \caption{Illustrative example of the two types of DG for a wireless multi-path channel estimation problem.}
    \label{fig:dg-types}
    \vspace{-0.2cm}
\end{figure*}

\noindent Fig. \ref{fig:dg-types} illustrates the difference between the vanilla and compound DG settings for estimating a wireless communication channel. There, we consider the wireless multi-path (MP) channel model $\bm{H}^{\textrm{MP}}(L,\,f)$ parametrized by the number of paths $L$ and the frequency $f$.
\noindent In vanilla DG, channel samples within the source dataset belong to \textit{known} domains, i.e., the number of paths $L$ is known for each sample. This is illustrated in Fig. \ref{fig:vanilla-dg} by clustering channel samples into three a priori known domains pertaining to three MP channel models associated with $L=1,5,15$. In compound DG, however, the domains of the channel samples are \textit{not known} as highlighted in Fig. \ref{fig:compound-dg}. Hence, the source dataset can be perceived as an unlabeled dataset where the domain knowledge of samples is unknown. In summary, prior knowledge of ``which samples belong to which domain'' is the main difference between vanilla DG different from compound DG.

Before delving into the details of DG algorithms, we now present the possible distribution shifts in the source/target domains and the related DG research fields. 

\subsection{Different Types of Domain Shifts}\label{subsec:types-distribution-shift}

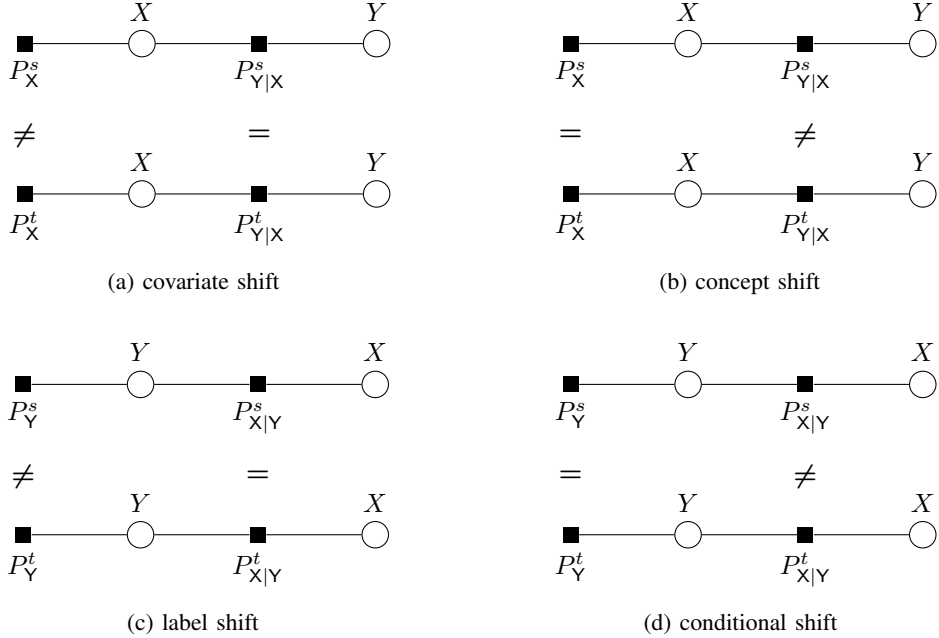
\begin{figure*}[t]
    \centering
    \subfloat[covariate shift \label{fig:covariate-shift}]{{
    \scalebox{1}{\begin{tikzpicture}[scale = 1.3]
    \newcommand{\vertex}{\node[vertex]}
    \tikzset{vertex/.style = {circle, draw, inner sep = 0pt, minimum size = 10pt}}
    \vertex[label = $X$](x1) at (-1.2,0) {};
    \vertex[label = $Y$](x2) at (1.2,0) {};
    \tikzset{vertex/.style = {rectangle, fill = black, inner sep = 0pt, minimum size = 6pt}}
    \vertex[label = below: $P_{\mathsf{X}}^t$](px) at (-2.4,0) {};
    \vertex[label = below: $P^t_{\mathsf{Y}|\mathsf{X}}$](delta) at (0,0) {};
    \draw (px)--(x1);
    \draw (delta)--(x1);
    \draw (delta)--(x2);
    \tikzset{vertex/.style = {circle, draw, inner sep = 0pt, minimum size = 10pt}}
    \vertex[label = $X$,yshift=2cm](x11) at (-1.2,0) {};
    \vertex[label = $Y$,yshift=2cm](x22) at (1.2,0) {};
    \tikzset{vertex/.style = {rectangle, fill = black, inner sep = 0pt, minimum size = 6pt}}
    \vertex[label = below: $P^s_{\mathsf{X}}$,yshift=2cm](pxx) at (-2.4,0) {};
    \vertex[label = below: $P^s_{\mathsf{Y}|\mathsf{X}}$,yshift=2cm](deltaa) at (0,0) {};
    \draw (pxx)--(x11);
    \draw (deltaa)--(x11);
    \draw (deltaa)--(x22);
    \node[] at ([yshift=0.6cm] px.center) {$\boldsymbol{\neq}$};
    \node[] at ([yshift=0.6cm] delta.center) {$\boldsymbol{=}$};
\end{tikzpicture}} 
    }}%
    \hspace{1.5cm}
    \subfloat[concept shift \label{fig:concept-shift}]{{
    \scalebox{1}{\begin{tikzpicture}[scale = 1.3]
    \newcommand{\vertex}{\node[vertex]}
    \tikzset{vertex/.style = {circle, draw, inner sep = 0pt, minimum size = 10pt}}
    \vertex[label = $X$](x1) at (-1.2,0) {};
    \vertex[label = $Y$](x2) at (1.2,0) {};
    \tikzset{vertex/.style = {rectangle, fill = black, inner sep = 0pt, minimum size = 6pt}}
    \vertex[label = below: $P^t_{\mathsf{X}}$](px) at (-2.4,0) {};
    \vertex[label = below: $P^t_{\mathsf{Y}|\mathsf{X}}$](delta) at (0,0) {};
    \draw (px)--(x1);
    \draw (delta)--(x1);
    \draw (delta)--(x2);
    \tikzset{vertex/.style = {circle, draw, inner sep = 0pt, minimum size = 10pt}}
    \vertex[label = $X$,yshift=2cm](x11) at (-1.2,0) {};
    \vertex[label = $Y$,yshift=2cm](x22) at (1.2,0) {};
    \tikzset{vertex/.style = {rectangle, fill = black, inner sep = 0pt, minimum size = 6pt}}
    \vertex[label = below: $P^s_{\mathsf{X}}$,yshift=2cm](pxx) at (-2.4,0) {};
    \vertex[label = below: $P^s_{\mathsf{Y}|\mathsf{X}}$,yshift=2cm](deltaa) at (0,0) {};
    \draw (pxx)--(x11);
    \draw (deltaa)--(x11);
    \draw (deltaa)--(x22);
    \node[] at ([yshift=0.6cm] px.center) {$\boldsymbol{=}$};
    \node[] at ([yshift=0.6cm] delta.center) {$\boldsymbol{\neq}$};
\end{tikzpicture}}
    }}%
    \vspace{0.5cm}
    \hspace{1cm}
    \subfloat[label shift \label{fig:label-shift}]{{
    \scalebox{1}{\begin{tikzpicture}[scale = 1.3]
    \newcommand{\vertex}{\node[vertex]}
    \tikzset{vertex/.style = {circle, draw, inner sep = 0pt, minimum size = 10pt}}
    \vertex[label = $Y$](x1) at (-1.2,0) {};
    \vertex[label = $X$](x2) at (1.2,0) {};
    \tikzset{vertex/.style = {rectangle, fill = black, inner sep = 0pt, minimum size = 6pt}}
    \vertex[label = below: $P^t_{\mathsf{Y}}$](px) at (-2.4,0) {};
    \vertex[label = below: $P^t_{\mathsf{X}|\mathsf{Y}}$](delta) at (0,0) {};
    \draw (px)--(x1);
    \draw (delta)--(x1);
    \draw (delta)--(x2);
    \tikzset{vertex/.style = {circle, draw, inner sep = 0pt, minimum size = 10pt}}
    \vertex[label = $Y$,yshift=2cm](x11) at (-1.2,0) {};
    \vertex[label = $X$,yshift=2cm](x22) at (1.2,0) {};
    \tikzset{vertex/.style = {rectangle, fill = black, inner sep = 0pt, minimum size = 6pt}}
    \vertex[label = below: $P^s_{\mathsf{Y}}$,yshift=2cm](pxx) at (-2.4,0) {};
    \vertex[label = below: $P^s_{\mathsf{X}|\mathsf{Y}}$,yshift=2cm](deltaa) at (0,0) {};
    \draw (pxx)--(x11);
    \draw (deltaa)--(x11);
    \draw (deltaa)--(x22);
    \node[] at ([yshift=0.6cm] px.center) {$\boldsymbol{\neq}$};
    \node[] at ([yshift=0.6cm] delta.center) {$\boldsymbol{=}$};
\end{tikzpicture}}
    }}%
    \hspace{1.5cm}
    \subfloat[conditional shift \label{fig:conditional-shift}]{{
    \scalebox{1}{\begin{tikzpicture}[scale = 1.3]
    \newcommand{\vertex}{\node[vertex]}
    \tikzset{vertex/.style = {circle, draw, inner sep = 0pt, minimum size = 10pt}}
    \vertex[label = $Y$](x1) at (-1.2,0) {};
    \vertex[label = $X$](x2) at (1.2,0) {};
    \tikzset{vertex/.style = {rectangle, fill = black, inner sep = 0pt, minimum size = 6pt}}
    \vertex[label = below: $P^t_{\mathsf{Y}}$](px) at (-2.4,0) {};
    \vertex[label = below: $P^t_{\mathsf{X}|\mathsf{Y}}$](delta) at (0,0) {};
    \draw (px)--(x1);
    \draw (delta)--(x1);
    \draw (delta)--(x2);
    \tikzset{vertex/.style = {circle, draw, inner sep = 0pt, minimum size = 10pt}}
    \vertex[label = $Y$,yshift=2cm](x11) at (-1.2,0) {};
    \vertex[label = $X$,yshift=2cm](x22) at (1.2,0) {};
    \tikzset{vertex/.style = {rectangle, fill = black, inner sep = 0pt, minimum size = 6pt}}
    \vertex[label = below: $P^s_{\mathsf{Y}}$,yshift=2cm](pxx) at (-2.4,0) {};
    \vertex[label = below: $P^s_{\mathsf{X}|\mathsf{Y}}$,yshift=2cm](deltaa) at (0,0) {};
    \draw (pxx)--(x11);
    \draw (deltaa)--(x11);
    \draw (deltaa)--(x22);
    \node[] at ([yshift=0.6cm] px.center) {$\boldsymbol{=}$};
    \node[] at ([yshift=0.6cm] delta.center) {$\boldsymbol{\neq}$};
\end{tikzpicture}}
    }}
    \caption{Factor graphs of the four possible distribution shifts in the source and target domain factorizations $P^s_{\mathsf{X},\mathsf{Y}}$ and $P^t_{\mathsf{X},\mathsf{Y}}$: (a) covariate shift where $P^s_{\mathsf{X}} \neq P^t_{\mathsf{X}}$, (b) concept shift where $P^s_{\mathsf{Y}|\mathsf{X}} \neq P^t_{\mathsf{Y}|\mathsf{X}}$, (c) label shift where $P^s_{\mathsf{Y}} \neq P^t_{\mathsf{Y}}$, and (d) conditional shift where $P^s_{\mathsf{X}|\mathsf{Y}}\neq P^t_{\mathsf{X}|\mathsf{Y}}$.}
    \label{fig:four-shifts}
    \vspace{-0.3cm}
\end{figure*}

Given a joint distribution $P_{\textsf{X},\textsf{Y}}$ associated with either a source or target domain, it is always possible to factorize it in two different forms using the Bayes rule:
\begin{subequations}\label{eq:distribution}
    \begin{align}
        P_{\textsf{X},\textsf{Y}}&= \underbrace{P_{\textsf{X}|\textsf{Y}}}_{\substack{\textrm{conditional}\\ \textrm{distribution}}}\,\times\, \underbrace{P_{\textsf{Y}}}_{\substack{\textrm{label}\\ \textrm{distribution}}}\hspace{-0.2cm},\label{eq:px-py/x}\\
         &= \underbrace{P_{\textsf{Y}|\textsf{X}}}_{\substack{\textrm{concept}\\ \textrm{distribution}}}\,\times\, \underbrace{P_{\textsf{X}}}_{\substack{\textrm{input}\\ \textrm{distribution}}}\hspace{-0.2cm}.\label{eq:py-px/y}
    \end{align}
\end{subequations}

\noindent  While the two domain factorizations in (\ref{eq:px-py/x}) and (\ref{eq:py-px/y}) are mathematically equivalent, they reveal two distinct direction of the causal relationship between the input random variable $\mathsf{X}$ and the label random variable $\mathsf{Y}$. By way of explanation, the factorization in (\ref{eq:px-py/x}) captures the causal relationship from $\mathsf{Y}$ to $\mathsf{X}$. This is because the conditional distribution of $\mathsf{X}$ given $\mathsf{Y}$ cannot be determined unless a particular realization of $\mathsf{Y}$ has been already observed to fully characterize the conditional distribution $P_{\textsf{X}|\textsf{Y}}$. Exchanging the roles of $\mathsf{Y}$ and $\mathsf{X}$ shows how the causal relationship from $\mathsf{X}$ to $\mathsf{Y}$ is captured by the factorization in (\ref{eq:py-px/y}). To see how this is the case, we resort in what follows to factors graphs as pictorial representations of joint probability distributions to efficiently communicating the conditional dependence structure between $\mathsf{X}$ and $\mathsf{Y}$ \cite{DBLP:journals/tit/KschischangFL01}. As depicted in Fig. \ref{fig:four-shifts}, random variables correspond to ``variables nodes'' represented in circles while distributions are associated with ``factor nodes'' illustrated in squares. Each variable node is connected to a factor node through an edge only when the factor node is dependent of the variable node.

In general, distribution shifts can influence at least one of the four distributions involved in the domain factorization in (\ref{eq:distribution}). For this reason, we distinguish four types of distribution shifts between the source and target domains. Fig. \ref{fig:four-shifts} depicts the factor graphs associated with each of the following distribution shifts:
\begin{itemize}[leftmargin=*]
    \item a distribution shift between the source and target input distributions, i.e., $P_{\textsf{X}}^{s} \neq P_{\textsf{X}}^{t}$, as shown in Fig. \ref{fig:covariate-shift}. This shift is commonly called \textit{covariate shift} \cite{shimodaira2000improving} and is the most studied type of distribution shift in the literature.\vspace{0.05cm}
    \item a distribution shift between the source and target concept distributions, i.e., $P_{\textsf{Y}|\textsf{X}}^s \neq P_{\textsf{Y}|\textsf{X}}^t$, as shown in Fig. \ref{fig:concept-shift}. The concept shift is usually not examined in DG classification tasks because most of the prior work assumes that data samples have different labels in different domains.\vspace{0.05cm}
    \item a distribution shift between the source and target label distributions, i.e., $P_{\textsf{Y}}^s \neq P_{\textsf{Y}}^{t}$, as illustrated in Fig. \ref{fig:label-shift}. This is called \textit{label shift} and is common in ML datasets, e.g., class imbalance in classification tasks.\vspace{0.05cm}
    \item a distribution shift between the source and target conditional distributions, i.e., $P_{\textsf{X}|\textsf{Y}}^{s} \neq P_{\textsf{X}|\textsf{Y}}^{t}$, as depicted in Fig. \ref{fig:conditional-shift}. This shift is often considered unchanged (i.e., $P_{\textsf{X}|\textsf{Y}}^{s} = P_{\textsf{X}|\textsf{Y}}^{t}$) to ensure that the label random variable $\textsf{Y}$ causes the input random variable $\textsf{X}$ in the same way between the source and target domain.\vspace{0.1cm}
\end{itemize}

\noindent Note that each type of distribution shift is often studied independently, and the existing algorithms for DG assume that the other shifts are not present \cite{DBLP:journals/corr/abs-1812-11806}. It is worth mentioning that most proposed algorithms in the literature focus on the covariate shift only and are specialized in classification tasks.

DG is closely related to other generalization concepts such as: multi-task learning \cite{DBLP:journals/ml/Caruana97}, transfer learning \cite{DBLP:journals/pieee/ZhuangQDXZZXH21}, zero-shot learning \cite{DBLP:journals/tist/WangZYM19}, domain adaptation \cite{DBLP:journals/ijon/WangD18}, and test-time training \cite{DBLP:journals/corr/abs-1909-13231}. Figure \ref{fig:tutoTaxonomy} illustrates the taxonomy of these generalization concepts and we will subsequently elaborate further on their differences. 

\subsection{Related Research Fields}

When the source and target domains are assumed to be the same (i.e., $P_{\textsf{X},\textsf{Y}}^s = P_{\textsf{X},\textsf{Y}}^t$), the learned model is not exposed to any domain shift. This assumption is pervasive in wireless communication ML applications where training and test datasets are usually generated using the \textit{same} system model and/or assumed to originate from the same propagation environments\footnote{By propagation environments, we refer to all the different parameters that impact the signal propagation conditions like path loss, coherence time, blockages, etc.}. However, in practice, this assumption is often violated and the test domains are usually different from the training domain(s). Supervised and multi-task learning are two common learning techniques where no domain shift occurs.\vspace{0.15cm}

\noindent\textbf{Supervised learning} learns a mapping between inputs and outputs assuming that training and test samples are identically and independently distributed (i.i.d). Supervised learning considers a \emph{single} domain ($M=1$). Because supervised learning does not handle domain shifts, the source and target samples are i.i.d. drawn from the same joint distribution, i.e., $P^s_{\mathsf{XY}}=P^t_{\mathsf{XY}}$. This is different from the DG setting where the i.i.d assumption is violated since the source and target samples are drawn from different distributions.\vspace{0.15cm}

\noindent\textbf{Multi-task learning} trains a single model to simultaneously perform multiple tasks, i.e., $M>1$. For the rest of the paper, a task refers to a type of problem to be solved such as classification or regression. Different tasks result in different but related domains or datasets which enable learning shared representations between tasks. Note that each task is characterized by a specific joint distribution $P_{XY}^{s_i} \neq P_{XY}^{s_j}$ for $i\neq j$ with $ 1 \leq i,j\leq M$. The objective of multi-task learning is to learn a model that performs well on the source tasks, meaning that target and source domains are the same. This is to be opposed to DG which aims to generalize to unseen domains/tasks.

\begin{figure*}[th!]
     \centering
         \centering
         \includegraphics[scale=0.68]{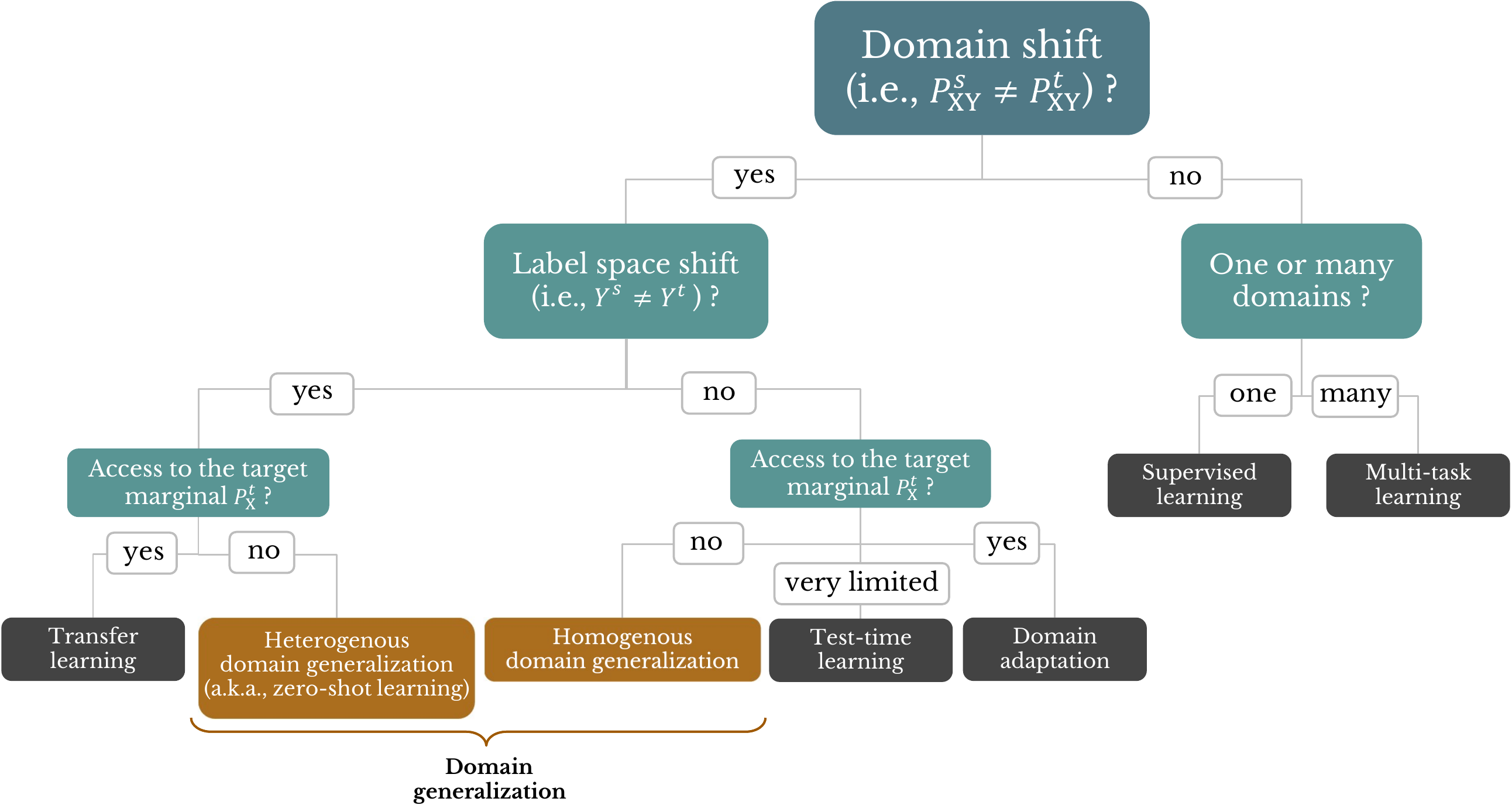}
         \caption{Similarities and differences between domain generalization covered in this work and other generalization related topics.}
         \label{fig:tutoTaxonomy}
\end{figure*}

When the source and target domains are not identical, i.e., $P_{\textsf{X},\textsf{Y}}^s \neq P_{\textsf{X},\textsf{Y}}^t$, the domain shift problem has to be addressed. In practice, DNNs trained on a given source domain suffer significant performance degradations on a different target domain even when the latter covers small variations compared to the source domain. In such cases, it is important to determine the origin and the assumptions behind the domain shift. As shown in Fig.~\ref{fig:tutoTaxonomy}, we overview four generalization paradigms in addition to DG. One important differentiating aspect between these paradigms is the \emph{access to the target data during training}. Domain adaptation and transfer learning both assume that test data is available which is not the case in DG and zero-shot learning.\vspace{0.15cm}

\noindent\textbf{Transfer learning} seeks to transfer the knowledge learned from one or multiple source domains/tasks to a different but related one. Finetuning is a common transfer learning technique where a model is first pre-trained on (often large) source datasets and then finetuned on different target datasets. The key difference between transfer learning and DG is access to the target data. Since transfer learning techniques involve model finetuning, target samples are required. However, DG assumes that target data is not accessible during training. In other words, transfer learning requires target data to generalize via additional training whereas DG seeks to achieve generalization on the target domain without any additional training. Nonetheless, both transfer learning and DG consider the target and source distributions to be different. Specifically, transfer learning usually deals with different label spaces which is also the case in heterogeneous DG.\vspace{0.15cm}

\noindent \textbf{Domain adaptation} considers the target marginal to be accessible at training time and hence can leverage target samples to improve the performance of DNN on target domains at inference time. The default setting of domain adaptation assumes that target samples are unlabelled and the target and source domain share the same label space (i.e., homogeneous domain adaptation). Different from domain adaptation, DG restricts the training to samples from the source domains only. As in the heterogeneous DG setting, heterogeneous domain adaptation extends the original setting by allowing for different label spaces between the source and target domains. Other domain adaptation variants are proposed in the literature such as zero-shot domain adaptation where target data are not required during training \cite{DBLP:conf/eccv/PengWE18}.\vspace{0.15cm}

\noindent \textbf{Zero-shot learning} primarily deals with label space shift (i.e., $\mathcal{Y}^s \neq \mathcal{Y}^t$). In other words, the goal of zero-shot learning is to recognize or generalize to classes or target values that are unseen during training. However, DG was initially proposed to handle the covariate shift arising from a change in the marginal distribution only (i.e., $P^s_X \neq P^t_X$). Note that zero-shot learning is related to heterogeneous DG since both covariate and label space shifts are allowed.\vspace{0.15cm}

\noindent \textbf{Test-time adaptation/training} aims to adapt a trained model to new domains without accessing the source data and human annotations at test time. Test-time training methods train a model to perform two tasks, namely the main task and the self-supervised auxiliary task. The self-supervised task will be used at test time to create labels for the unlabeled test samples. The standard version of test-time training requires a very limited amount (e.g., a mini-batch) of data to fine tune the model based on the auxiliary task. This is where test-time training differs from DG due to the use of test data for updating the model parameters.\vspace{0.15cm}

\noindent\textbf{Continual/lifelong learning} learns a model on multiple domains or tasks sequentially without forgetting the knowledge previously learned. Continual learning assumes that the model does not have access to data from previous tasks and updates the parameters using labeled data from new tasks or domains. This is different from DG where the objective is to generalize to new domains without accessing target data or finetuning the model on the target domain.\vspace{0.1cm}

\begin{figure}[!ht]
     \centering
         \centering
         \includegraphics[scale=0.48]{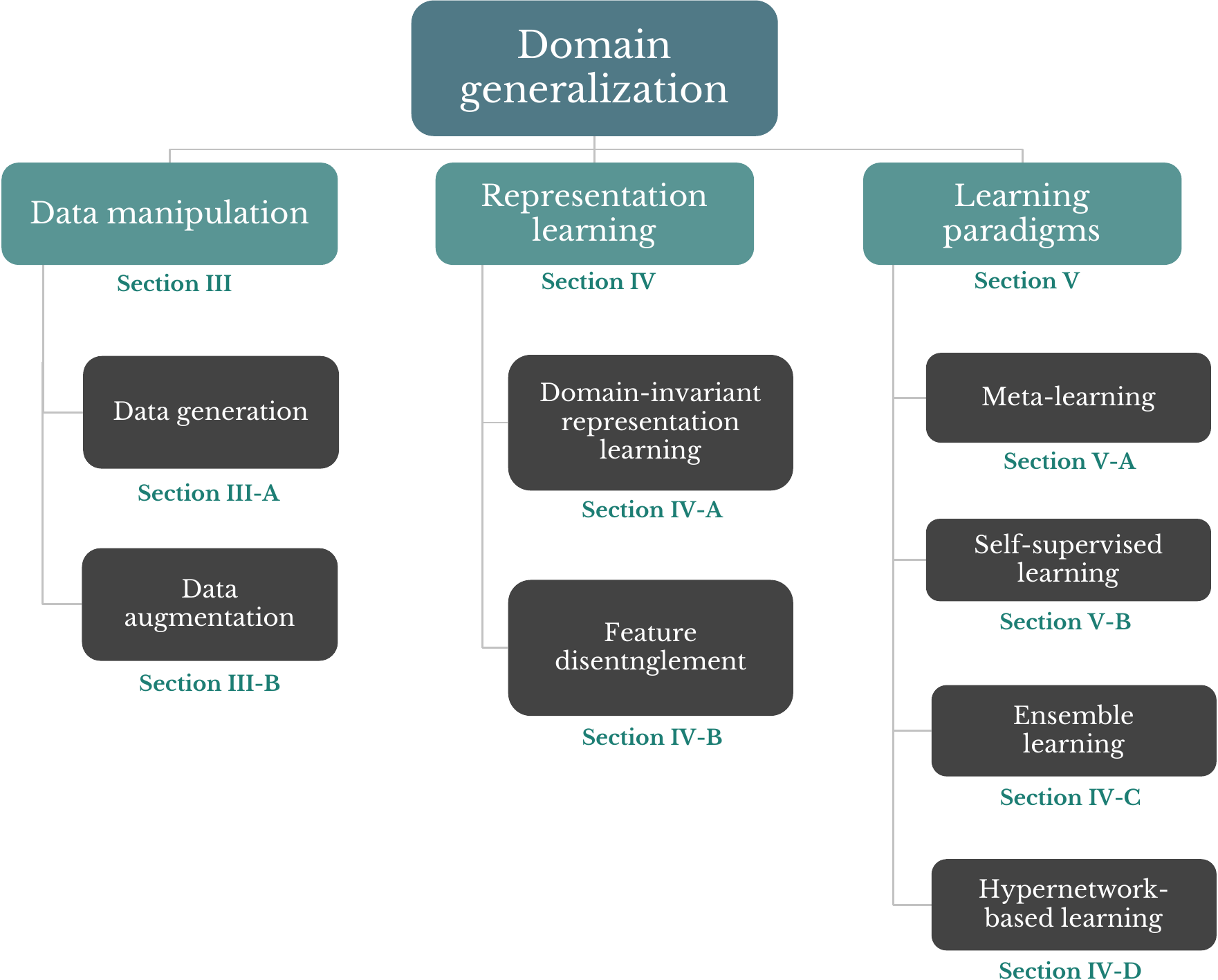}
         \caption{Taxonomy of domain generalization methods.}
         \label{fig:DGTaxonomy}
\end{figure}

In the next sections, we put forward the current state-of-the-art methodologies for handling DG. We also specify which methods cope with distribution shifts beyond the covariate shift. Fig. \ref{fig:DGTaxonomy} presents the organization of the covered DG methodologies across the next three sections.

\section{DG Methods: Data Manipulation}\label{sec:data-manipulation}
In order to generalize to unseen scenarios, this category of methods manipulates the DNN input data. Two types of manipulations are possible either in the raw input space or in the latent input space: $i)$ data augmentation by adding random noise or transformation to the input data, and $ii)$ data generation which generates new training samples using generative models. The main objective of these methods is to increase the quantity and improve the diversity of the training dataset for better generalization capabilities without requiring manual labeling of datasets.

\noindent A data manipulation operation is represented by an arbitrary function $\mathcal{M}(\cdot)$ which transforms the input data $X$ to the manipulated data $X^\prime = \mathcal{M}(X)$. Given a DNN that is represented as an input-output function $g(\cdot)$, the learning objective of data manipulation for DG can be expressed as follows:
\begin{equation}\label{eq:data-manipulation-cost-function}
\min_{g}\,\underbrace{\mathbb{E}_{{\mathsf{X}},{\mathsf{Y}}}\big[\mathcal{L}(g(X),Y)\big]}_{\textrm{task loss}} + \underbrace{\mathbb{E}_{\mathsf{X}^{'},\mathsf{Y}}\big[\mathcal{L}(g(X^\prime),Y)\big]}_{\textrm{data manipulation loss}},
\end{equation}
where $\mathcal{L}(\cdot,\cdot)$ is the DNN cost function. It is worth noting that most data manipulation techniques proposed in the literature are geared towards computer vision applications where all datasets consist of images. In this section, we describe these methods within the context of vision applications and point out their potential use for wireless applications.

\subsection{Data Generation}
Generating new data samples using generative models is a popular technique to augment existing datasets so as to cover richer training scenarios, thereby enhancing the generalization capability of a DNN. The data manipulation function $\mathcal{M}(\cdot)$ in (\ref{eq:data-manipulation-cost-function}) can be represented by deep generative models such as variational auto-encoder (VAE) \cite{DBLP:journals/corr/KingmaW13} and generative adversarial network
(GAN) \cite{DBLP:journals/corr/GoodfellowPMXWOCB14}.

Various distribution distance metrics can be employed to generate high-quality samples including:
\begin{itemize}
    \item \textit{domain discrepancy measures} such as the maximum mean discrepancy (MMD) \cite{DBLP:journals/jmlr/GrettonBRSS12} to minimize the distribution divergence between real and generated data samples.
    \item \textit{the Wasserstein distance} between the prior distribution of the DNN input and a latent target distribution as carried out in Wasserstein auto-encoder (WAE) \cite{DBLP:journals/corr/abs-1711-01558}. This metric is a regularization that encourages the encoded training distribution of a WAE to match the data prior and hence preserves the semantic and domain transfer capabilities.
    \item \textit{semantic consistency loss functions} that maximize the difference between the source and the newly generated distributions, thereby creating new domains that augment the existing source domains \cite{DBLP:conf/eccv/ZhouYHX20}.
\end{itemize}

\noindent It is also possible to generate new domains instead of new data samples using adversarial training \cite{DBLP:conf/cvpr/LiGCHWMYLX21} where one or multiple generative models are trained to progressively generate unseen domains by learning relevant cross-domain invariant representations. Such an alternative involves an entire generative model pipeline composed of multiple DNNs trained in cascade or in parallel, and therefore has a significant computational cost. As one example for channel estimation problems, one can start by generating line-of-sight datasets and then progressively increase the rank of the estimated MIMO channel to multi-path models up to full-rank channels such as rich-scattering MIMO channels.

Furthermore, the data manipulation function $\mathcal{M}(\cdot)$ can also be defined without training generative models. In particular, it is possible to generate new data samples by linearly interpolating any two training samples and their associated labels as done in the low-complexity Mixup method \cite{DBLP:conf/iclr/ZhangCDL18}. More recently, many techniques have built upon Mixup for DG to $i)$ generate new data samples by interpolating either in the raw data space \cite{DBLP:conf/bmvc/WangL0K021,DBLP:conf/icassp/WangLK20,DBLP:conf/cvpr/ShuCW0L21}, or $ii)$ to build robust models with better generalisation capabilities by interpolating in the feature space \cite{DBLP:conf/iclr/ZhouY0X21,DBLP:conf/cvpr/Qiao021,DBLP:conf/cvpr/XuZ0W021}.

\subsection{Data Augmentation}

DNNs are heavily reliant on large datasets to enhance the generalization by avoiding overfitting \cite{DBLP:journals/jbd/ShortenK19}. Data augmentation methods provide a cheap way to augment training datasets. They artificially inflate the dataset size by transforming existing data samples while preserving labels. Data augmentation includes geometric and color transformations for visual tasks, random erasing and/or permutation, adversarial training, and neural style transfer. Every data augmentation operation can be considered as a data manipulation function $\mathcal{M}(\cdot)$ in (\ref{eq:data-manipulation-cost-function}). Here, we classify the data augmentation methods for DG into two categories:
\begin{itemize}
    \item \textit{domain randomization}: this family of methods creates a variety of datasets stemming from data generation processes (e.g., simulated environments) with randomized properties and trains a model that generalizes well across all of them.
    \item \textit{adversarial data augmentation}: this family of methods guides the augmentation by enhancing the diversity of the dataset while ensuring their reliability for better generalization capabilities.
\end{itemize}

\begin{figure*}[ht]
     \centering
         \centering
         \includegraphics[scale=0.65]{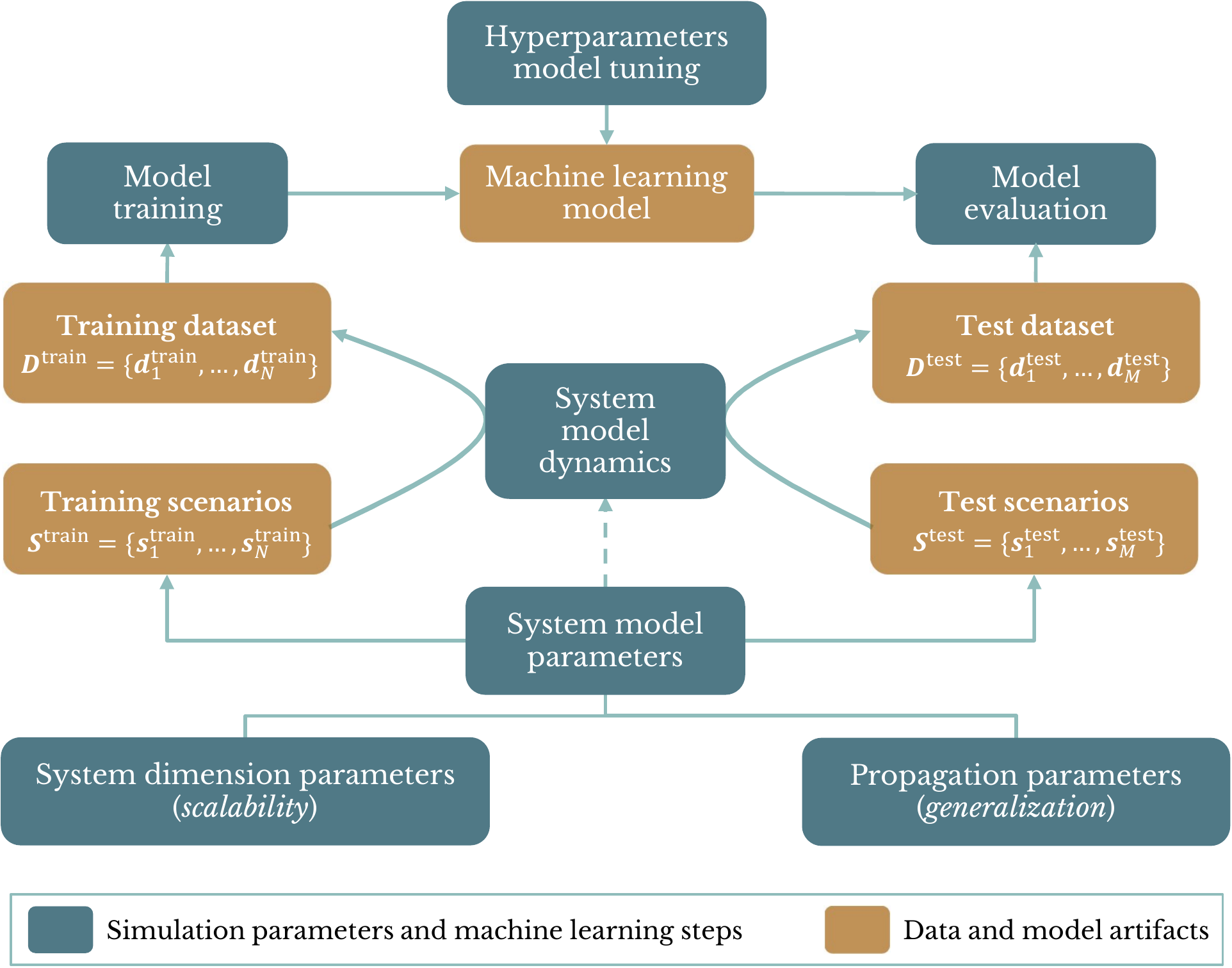}
         \vspace{0.15cm}
         \caption{Summary of the training and evaluation pipeline of machine learning models under data distribution shifts for communication applications.}
         \label{fig:domain-randomization}
\end{figure*}

\subsubsection{\textbf{Domain randomization}}\label{subsubsec:domain-randomization} The reality gap between the data domains resulting from simulations and real-world data collections often leads to failure due to distribution shifts. This gap is triggered by an inconsistency between the physical parameters of simulations (e.g., channel distribution, noise level) and, more fatally, the incorrect physical modeling (e.g., physical considerations of wireless communication \cite{DBLP:journals/tcas/IvrlacN10,akrout2022achievable}). To perceive how DNNs should be trained and evaluated under data distribution shifts for communication applications, Fig.~\ref{fig:domain-randomization} depicts the training and evaluation pipeline where datasets are generated through communication systems models. There, it is seen that source (i.e., training) and target (i.e., test) domains, $\mathcal{D}^{\textrm{train}}$ and $\mathcal{D}^{\textrm{test}}$, are obtained according to the training and test scenarios, $\mathcal{S}^{\textrm{train}}$ and $\mathcal{S}^{\textrm{test}}$. The latter are determined by defining a set of communication scenarios by varying one or multiple communication parameters of interest. The choice of these parameters dictates the data domains and hence provides a way to control and then analyze the impact of distribution shifts on the performance of DNNs. For instance, research efforts to design broadband ML-aided decoding algorithms should vary the signal frequency and assess the generalization capability of DNNs when trained on carriers in the sub-6 GHz band then evaluated on a different communication band. 

\noindent Domain randomization generates new data samples stemming from simulated dynamics of complex environments. For computer vision applications, the function $\mathcal{M}(\cdot)$ in (\ref{eq:data-manipulation-cost-function}) encloses different manual transformations such as altering object properties (e.g., shape, location, texture), scene editing (e.g., illumination, camera view), or random noise injection \cite{DBLP:conf/iros/TobinFRSZA17}. For real-valued data input vectors, augmentation involves scaling, pattern switching, and random perturbation \cite{pialla2022data}. These augmentation methods are particularly interesting for wireless communication applications because they handle general signal transmission scenarios that are tolerant to variations in the path-loss coefficient, synchronization delays, signal-to-noise ratio, etc.

\subsubsection{\textbf{Adversarial data augmentation}}
\noindent The fact that most domain randomization described in Section \ref{subsubsec:domain-randomization} is performed randomly indicates that there exist potential improvements to remove ineffective randomization that does not help with DNNs' generalization. This optimization is performed by adversarial data augmentation.

Toward this goal, research efforts have been dedicated to designing better strategies for non-random data augmentation. By modeling the dependence between the data sample $X$, its label $Y$, and the domain label $d$ (cf. Definition \ref{def:DG}), it has been shown that the input data can be perturbed along the direction of greatest domain change (i.e., domain gradient) while changing the class label as little as possible \cite{DBLP:conf/iclr/ShankarPCCJS18}. Another line of work devised an adaptive data augmentation procedure where adversarially perturbed samples in the feature space are iteratively added to the training dataset \cite{DBLP:conf/nips/VolpiNSDMS18}. It is also possible to train a dedicated transformation network for data augmentation by $i)$ maximizing the domain classification loss on the transformed data samples to tolerate domain generation differences, and $ii)$ minimizing the label classification loss to ensure that the learned augmentation does not affect the DNN performance \cite{DBLP:conf/aaai/ZhouYHX20}. While adversarial data augmentation can provide richer datasets and fill in data gaps against some adversarial examples, this comes at the cost of a more complex training procedure which is known to be less stable and computationally extensive.

When it comes to wireless communications applications, physics-based models are available to guide data augmentations that are consistent with the law of physics, beyond purely random strategies. For example, the study of the achievable rate of reconfigurable intelligent surface (RIS)-aided communication systems do exhibit the same performance regardless of the carrier frequency due to the scaling invariance property of Maxwell's equations when no source is present (i.e., passive RISs) \cite{jackson1999classical}. Another interesting implication stemming from the symmetry of Maxwell's equations is the frequency independence property of certain wideband antennas that display very similar radiation pattern, gain and impedance above a certain threshold frequency \cite{hohlfeld1999self}. This suggests that the generation of wireless datasets for far-field communication can be made independent of the carrier frequency for specific types of antennas.

From this perspective, data augmentation methods that are aware of the physics of wave propagation do not blindly generate source and target domains for different carrier frequencies. They should instead collapse the data augmentation process to scenarios that do enjoy the scaling invariance property. As a result, not only do data augmentation techniques become efficient but also physically consistent with the electromagnetic properties of RISs. 

\section{DG Methods: Representation Learning}\label{sec:representation-learning}
\begin{figure}[th!]
     \centering
     \begin{subfigure}[b]{0.49\textwidth}
         \centering
         \includegraphics[scale=0.32]{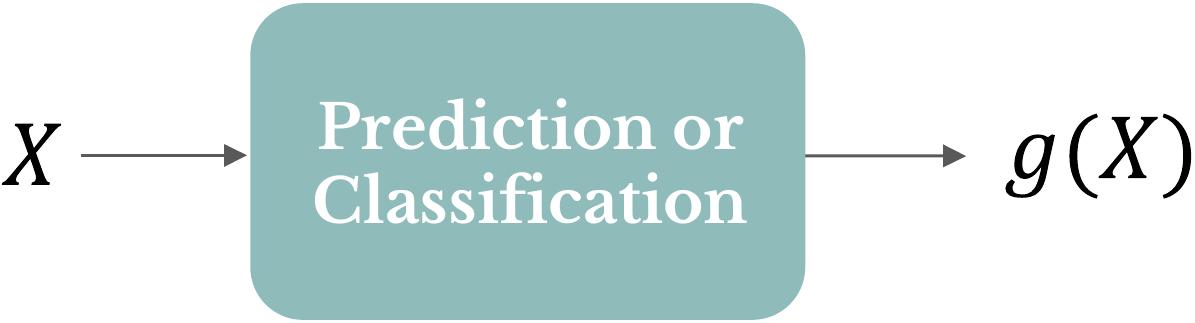}
         \caption{without representation learning\vspace{0.5cm}}
         \label{fig:DL-standard}
     \end{subfigure}
     
     \begin{subfigure}[b]{0.49\textwidth}
         \centering
         \includegraphics[scale=0.32]{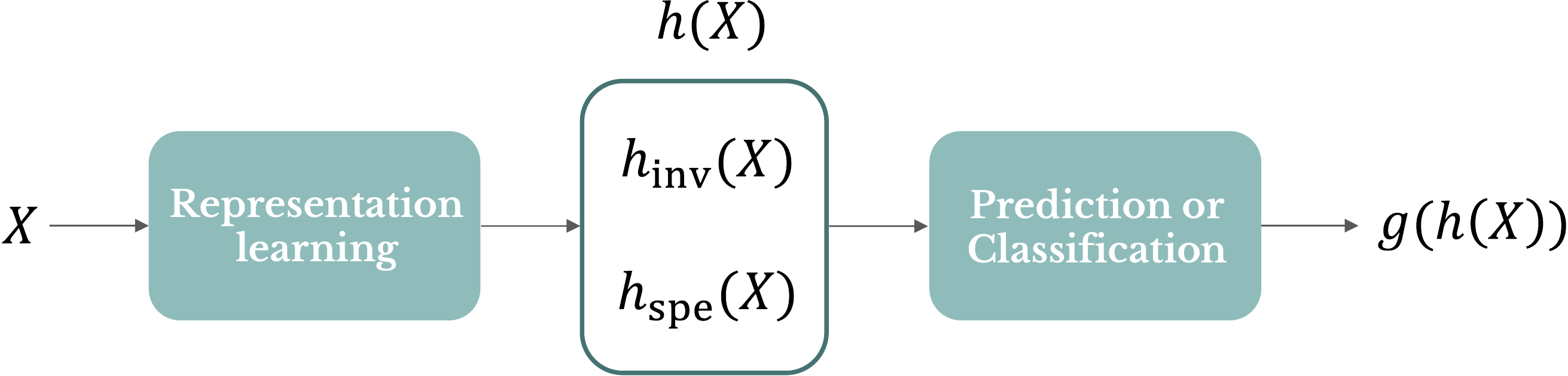}
         \caption{with representation learning}
         \label{fig:DL-rep-learning}
     \end{subfigure}
    \caption{Illustration of ML-aided classification/prediction (a) without an explicit representation learning step (a.k.a. end-to-end learning), and (b) with a representation learning step.}
    \label{fig:with-without-rep-learning}
\end{figure}

Generalizing to unseen scenarios is not solely dependent on the DNN prediction approximation function $g(\cdot)$ given in (\ref{eq:data-manipulation-cost-function}). It also depends on the data representations (i.e., features) learned by the DNN \cite{DBLP:journals/pami/BengioCV13}. To better isolate these two distinct tasks, one can view the overall DNN approximation function, $g(h(\cdot))$, as a composition of a prediction/classification function $g(\cdot)$ and a representation learning function $h(\cdot)$. Fig. \ref{fig:DL-rep-learning} depicts this decomposition, $h(X)$, as the output of the representation learning step. In theory, this representation in the feature space comprises two separate representations. The first one denoted by $h_{\textrm{inv}}(X)$ is a domain-invariant representation that is shared across domains (a.k.a., cross-domain representation) and is key to enabling generalization over multiple domains. The second representation $h_{\textrm{spe}}(X)$, however, is domain-specific and represents the variation pertaining to a specific domain. In practice, these two representations can either be non-separable or separable. For instance, several earlier research works \cite{DBLP:conf/icassp/OppenheimLKP79,oppenheim1981importance,hansen2007structural} have shown that in the Fourier spectrum of signals, the phase component predominantly carries low-level statistics whereas the amplitude component mainly contains high-level semantics. Hence, Fourier phase features represent domain-invariant features that cannot be easily affected by domain shifts when used for DG \cite{lu2022domaininvariant}.

From a mathematical point of view, the optimization problem of representation learning can be written as follows:

\begin{equation}\label{eq:representation-learning-cost-function}
\min_{g,\,h}~\,\underbrace{\mathbb{E}_{{\mathsf{X}},{\mathsf{Y}}}\big[\mathcal{L}(g(h(X)),Y)\big]}_{\textrm{task loss}} + ~\lambda\,\hspace{-0.5cm}\underbrace{r(X)}_{\textrm{regularization loss}},
\end{equation}
where $r(X)$ is a regularization function and $\lambda$ is the associated regularization parameter.

\noindent Depending on the type of the regularization function $r(X)$ or the representation learning function $h(\cdot)$, it is possible to categorize representation learning for DG into two categories:
\begin{itemize}
    \item \textit{domain-invariant representation learning}: the goal of this family of methods is to learn features that are invariant across different domains. These features are transferable from one domain to another, hence their importance for domain generalization.
    
    \item \textit{feature disentanglement}: these methods decompose a feature representation into one or multiple sub-features, each of which is either domain-specific or domain-invariant.
\end{itemize}

\subsection{Data-Invariant Representation Learning}\label{subsec:data-invariant-representation-learning}

\subsubsection{\textbf{Kernel-based methods}}
Learning representation using kernel methods (e.g., support vector machines \cite{DBLP:journals/ml/CortesV95}, kernel component analysis \cite{scholkopf1997kernel}) is a classical problem in the ML literature. In such a setting, the representation learning function $h(\cdot)$ in (\ref{eq:representation-learning-cost-function}) maps the data samples to the feature space using kernel functions (e.g., radial basis function (RBF), Gaussian, and Laplacian kernels).

For domain generalization, several methods were devised to learn domain-invariant kernels to determine $h(\cdot)$ from the training dataset. Specifically, a positive semi-definite kernel learning approach for DG was proposed in \cite{DBLP:journals/jmlr/BlanchardDDLS21} by considering the conventional supervised learning problem where the original feature space is augmented to include the marginal distribution that generates the features. It is also possible to learn kernel functions by minimizing the distribution discrepancy between all the data samples in the feature space. This method is known as domain-invariant component analysis (DICA) \cite{DBLP:conf/icml/MuandetBS13} and is one of the classical kernel methods for DG.

For classification tasks, in presence of covariate shift only, a randomized kernel algorithm was devised in \cite{DBLP:conf/ijcai/ErfaniBMNLBR16} to extract features that minimize the difference between the marginal distributions across domains. Multi-domain discriminant analysis (MDA) and scatter component analysis (SCA) approaches were proposed in \cite{DBLP:conf/uai/Hu0CC19,DBLP:journals/pami/GhifaryBKZ17} to learn a domain-invariant feature transformation in presence of both covariate and conditional shifts across domains. This is done by jointly minimizing the divergence among domains within each class and maximizing the separability among classes.

\subsubsection{\textbf{Domain adversarial learning}}

Since the presence of spurious features in the data decreases the robustness of DNNs, adversarial learning is a widely used technique to learn invariant features by training generative adversarial networks (GANs). Specifically, the discriminator is trained to distinguish the domains while the generator is trained to fool the discriminator so as to learn domain invariant feature representations for DG \cite{DBLP:conf/cvpr/LiPWK18}. Another line of work in \cite{DBLP:conf/cvpr/GongLCG19} generated a continuous sequence of intermediate domains flowing from one domain to another to gradually reduce the domain discrepancy, and hence improve the DNN generalization ability on unseen target domains. Learning class-wise adversarial networks for DG was also  proposed in \cite{DBLP:conf/eccv/LiTGLLZT18} based on conditional invariant adversarial training when both covariate and conditional shifts coexist.

\subsubsection{Explicit feature alignment}\label{subsubsec:explicit-feature-alignment}

This family of methods learns domain-invariant representations by aligning the features across source domains using one of the following two mechanisms:
\begin{itemize}
    \item explicit feature distribution alignment through distance minimization or moment matching.
    \item feature normalization addressing data variations to avoid learning nonessential domain-specific features.
\end{itemize}

Feature distribution alignment methods were devised to impose a variety of distribution distances such as the maximum mean discrepancy (MMD) on latent feature distributions \cite{DBLP:conf/cvpr/LiPWK18,DBLP:journals/corr/TzengHZSD14}, and the label similarities for samples of the same classes from different domains using the Wasserstein distance \cite{DBLP:journals/ijon/ZhouJSWC21}. Moment matching for multi-source domain adaptation (M3SDA) was also introduced in \cite{DBLP:conf/iccv/PengBXHSW19} to transfer learned features from multiple labeled source domains to an unlabeled target domain by dynamically aligning moments of their feature distributions.

Feature normalization methods, however, focus on increasing the discrimination capability of DNNs. They do so by normalizing the features to eliminate domain-specific variation while keeping domain-invariant features to enhance generalization. In particular, instance normalization (IN) \cite{DBLP:conf/eccv/PanLST18} and batch instance normalization (BIN) \cite{DBLP:conf/nips/NamK18} have been proposed to enhance the generalization capabilities of convolutional neural networks (CNNs). Instance normalization has been applied in \cite{DBLP:journals/corr/abs-2111-15077} for DG where labels were missing in the training domains to acquire invariant and transferable features. It was also shown that adaptively learning the normalization technique can improve DG without predefining the normalization technique in the DNN architecture a priori \cite{DBLP:conf/cvpr/FanWKYGZ21}.

\subsubsection{\textbf{Invariant risk minimization}}
Another unique perspective on learning domain-invariant representations for DG is to constrain DNNs to have the same output across all domains. The motivation behind this constraint is that an optimal representation for prediction or classification is \textit{the cause} of the DNN output label. This causal relationship from the representation (i.e., the cause) to the label (i.e., the effect) should not be affected by other factors including the domain input. Therefore, the optimal representation is domain invariant and can be learned using invariant risk minimization (IRM) \cite{DBLP:journals/corr/abs-1907-02893}. Given $K$ different domains, the IRM problem can be formulated as follows:
\begin{subequations}\label{eq:IRM}
    \begin{align}
    &\min_{h\in\mathcal{H}}~ \sum\limits_{k=1}^{K}~\mathbb{E}_{\,\mathsf{X}_k,{\mathsf{Y}}_k}\big[\mathcal{L}(g(h(X_k)),Y_k)\big]\label{eq:IRM-1}\\
    &\textrm{subject to~} g \in \bigcap_{k=1}^{K}\,\argmin_{g'\in\,\mathcal{G}}~\mathbb{E}_{\,{\mathsf{X}}_k,{\mathsf{Y}}_k}\big[\mathcal{L}(g'(h(X_k)),Y_k)\big],\label{eq:IRM-2}
    \end{align}
\end{subequations}
where $\mathcal{H}$ and $\mathcal{G}$ are the learnable function classes for representation and task functions, $h(\cdot)$ and $g(\cdot)$, respectively. The optimization in (\ref{eq:IRM}) finds the optimal representation function $h(\cdot)$ that minimizes the sum of all the task losses in (\ref{eq:IRM-1}) given in (\ref{eq:representation-learning-cost-function}). This minimization is carried out under the constraint in (\ref{eq:IRM-2}) which ensures that all domains share
the same optimal representation function $h(\cdot)$.

The idea behind the IRM formulation has drawn significant attention. Specifically, the IRM optimization was extended to text classification \cite{DBLP:journals/corr/abs-2004-05007}, reinforcement learning \cite{DBLP:conf/l4dc/SonarPM21}, self-supervised settings \cite{DBLP:conf/iclr/MitrovicMWBB21}, and to the case of extrapolated task losses among source domains \cite{DBLP:conf/icml/KruegerCJ0BZPC21}. Moreover, it was shown in \cite{DBLP:conf/nips/AhujaCZGBMR21} that constraining the invariance to the task function $g(\cdot)$ only~--- as done in (\ref{eq:IRM})~--- is not enough to guarantee the causal relationship from the representation to the label. A new regularization has thus been proposed to ensure that the representation function $h(\cdot)$ cannot capture fully invariant features that break down the assumed causality as required by the IRM formulation.

\subsection{Feature Disentanglement}

\begin{figure}[h!]
     \centering
         \centering
         \includegraphics[scale=0.48]{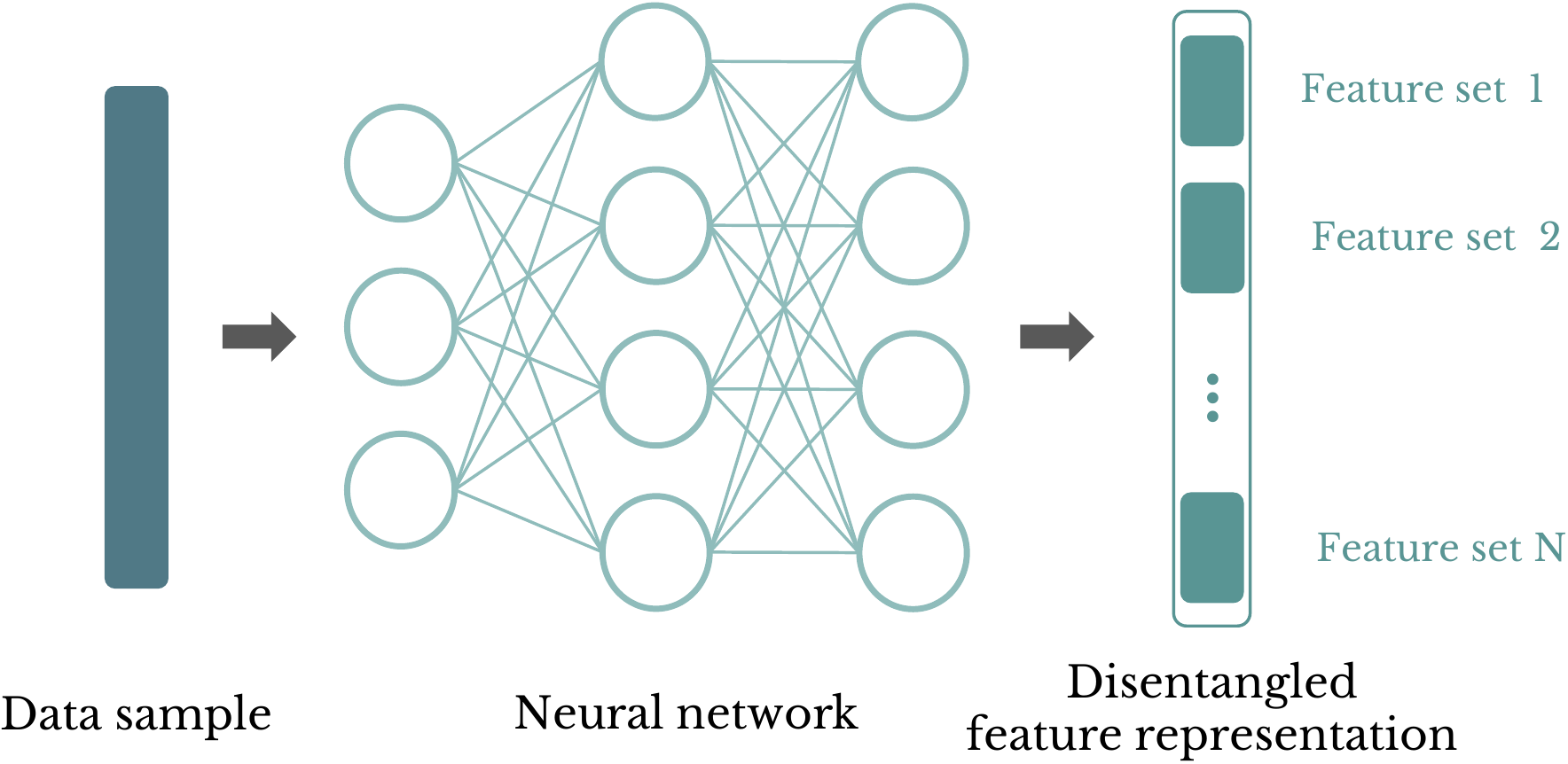}
         \caption{Illustration of how a trained neural network transforms a data sample into a disentangled representation vector that factorizes into $N$ small feature vectors.}
         \label{fig:disentangled-representation}
\end{figure}

Unlike domain-invariant representation learning, disentangled representation learning relies on DNNs to learn a function that maps a data sample to a feature vector, which factorizes into distinct feature sets as depicted in Fig. \ref{fig:disentangled-representation}. There, it is seen that the entire feature space can be decomposed into a set of feature subspaces. Each feature set is a representation pertaining to a specific feature subspace only. When the feature representation is decomposable into multiple non-overlapping feature subsets, the feature representation is said to be ``disentangled''.

\noindent The importance of disentanglement-based representation learning for DG stems from the fact that features can be explicitly decomposed into domain-invariant and domain-specific features. As a result, the representation function $h(\cdot)$ defined in (\ref{eq:representation-learning-cost-function}) can be decomposed into two distinct representation functions: $h_{\textrm{inv}}(\cdot)$ for domain-invariant representation and $h_{\textrm{spe}}(\cdot)$ for domain-specific representation. The disentanglement-based optimization can be formulated as follows:
\begin{equation}\label{eq:disentangled-representation-learning-cost-function}
\begin{aligned}
\hspace{-0.3cm}\min_{h_{\textrm{spe}},\,h_{\textrm{inv}},\,g}~&\,\underbrace{\mathbb{E}_{{\mathsf{X}},{\mathsf{Y}}}\big[\mathcal{L}(g(h_{\textrm{inv}}(X)),Y)\big]}_{\textrm{task loss}} +~ \lambda\,\hspace{-0.5cm}\underbrace{r(X)}_{\textrm{regularization loss}} \\
&\hspace{1cm}+ \mu\,\,\underbrace{\mathbb{E}_{{\mathsf{X}}}\big[\mathcal{L}(h_{\textrm{inv}}(X),h_{\textrm{spe}}(X),X)\big]}_{\textrm{reconstruction loss}},
\end{aligned}
\end{equation}
where $\lambda$ and $\mu$ are regularization parameters. In (\ref{eq:representation-learning-cost-function}), the regularization loss encourages the separation between domain-invariant and domain-specific features, while the reconstruction loss ensures that such separation does not lead to significant information loss. In other words, regularization and reconstruction losses are competing penalties that add up to the task loss, and it is the task of the ML designer to find the suitable trade-off that enhances the generalization of DNNs.

\subsubsection{\textbf{Multi-component analysis}}
Multi-component methods dedicate different sets of parameters to learn domain-invariant and domain-specific features. The method ``UndoBias'' proposed in \cite{DBLP:conf/eccv/KhoslaZMET12} learns dedicated SVM models. It represents the dedicated SVM parameters, $\bm{w}_k$, pertaining to the $k$th domain as a perturbation of the domain-invariant parameters $\bm{w}$ with the domain-specific parameters $\Delta\bm{w}_k$, i.e., $\bm{w}_k = \bm{w} + \Delta\bm{w}_k$. This method has been extended for multi-view vision tasks by introducing a regularization to minimize the mismatch between any two view representations \cite{DBLP:conf/iccv/NiuLX15} for better generalization. Neural networks have also been used to capture disentangled representations by learning domain-specific
networks for each domain and one domain-invariant network for all domains \cite{DBLP:journals/tip/DingF18}. Another line of work considered manually comparing specific areas of DNN's attention heatmaps from different domains which proved beneficial to learning disentangled representations and ensuring a more robust generalization \cite{DBLP:conf/cvpr/ZuninoBVS0SMS21}.

\subsubsection{\textbf{Generative modeling}}
Generating data samples whose feature representations are disentangled requires adapting the data generative process of generative models to new constraints. The latter can be incorporated in the loss functions of GANs to encourage feature disentanglement
by separating the domain-specific and domain-invariant features \cite{DBLP:journals/corr/abs-2109-05826}. An autoencoder-based variational approach was devised to disentangle the features by learning three independent latent subspaces, one for the domain, one for the class, and one for any residual variations \cite{DBLP:conf/midl/IlseTLW20}. To generate domains that are different from the source domain, the discrepancy between augmented and sources domains was maximized for out-of-domain augmentation using meta-learning under a semantic consistency constraint \cite{DBLP:conf/cvpr/QiaoZP20}.

For classification tasks, diversifying the inter-class
variation by modeling potential seen or unseen variations across classes was formulated as a disentanglement-constrained optimization problem \cite{DBLP:conf/cvpr/ZhangZLWSX22}. This was made possible by minimizing the discrepancy of the inter-class variation where both intra- and inter-domain variations are regarded as constraints.

\section{DG Methods: Learning Paradigms}\label{sec:learning-paradigms}
\subsection{Meta-learning}\label{subsec:meta-learning}
Meta-learning \cite{DBLP:journals/air/HuismanRP21}, also known as learning-to-learn is a research area that has attracted much interest in recent years. The main goal of meta-learning is to learn a general model using samples from multiple tasks to quickly adapt to new unseen tasks. The learned meta-model encompasses the general knowledge from all the different training tasks which makes it a better model initialization to adapt for new tasks \cite{DBLP:conf/icml/FinnAL17}. 

Traditional supervised learning (SL) methods learn a model $f_\theta$ that maps inputs to outputs. The model's parameters $\theta$ are learned by minimizing a loss function given a dataset $\mathcal{D} = \{(x_i,y_i)\}_{i=1}^{m}$ as follows:
\begin{align*}
    \theta^*_{\textrm{SL}} = \argmin_\theta \mathcal{L}(\mathcal{D}, \theta)
\end{align*}
At each iteration, the parameters are updated based on a specific optimization procedure $g_\omega$ where $\omega$ denotes all the pre-defined assumptions about the learning algorithm such as the function class of $f$ (e.g., DNN), the initial model initialization, the choice of the optimizer, etc. In the literature, $\omega$ is also called \emph{pre-defined meta-knowledge} \cite{DBLP:journals/pami/HospedalesAMS22}. It is straightforward to observe that the model's performance depends drastically on $\omega$. 

In addition, it is common to split the dataset $\mathcal{D}$ into training and testing sets. The model is first learned using the training samples, and the generalization of the model is subsequently evaluated on the test set with unseen samples and known outputs. Consequently, the learned parameters $\theta_{\textrm{SL}}$ are specific to the dataset $\mathcal{D}$ and are not guaranteed to generalize to samples different from the ones in $\mathcal{D}$. 

Different from the supervised learning setting, meta-learning aims to learn a meta-knowledge $\omega$ over a distribution of tasks $p(\mathcal{T})$. A task $i$ can be defined by a loss function and a dataset (i.e., $\mathcal{T}_i=\{\mathcal{L}_i, \mathcal{D}_i\}$). Learning the meta-knowledge from multiple tasks enables the quick learning of new tasks from $p(\mathcal{T})$. In meta-learning, different choices of the meta-knowledge $\omega$ are proposed such as parameter initialization, optimizer, hyperparameters, task-loss functions, etc. We refer the interested reader to \cite{DBLP:journals/pami/HospedalesAMS22} for a detailed discussion about the different choices for $\omega$.

Meta-learning algorithms also involve two stages, namely meta-training followed by meta-testing. The objective of meta-training is to learn the ``best'' meta-knowledge $\omega$ across multiple tasks. To do so, a set of training tasks $\mathcal{T}_{\textrm{train}} \sim p(\mathcal{T})$ is used where each task $i$ has training and validation datasets (i.e., $\mathcal{D}_i=\{\mathcal{D}_i^{\textrm{train}}, \mathcal{D}_i^{\textrm{val}}\}$). The meta-training phase is commonly presented as a bi-level optimization problem \cite{DBLP:journals/pami/HospedalesAMS22} as follows:
\begin{align}
   \omega^{*} &= \overbrace{\mathop{\argmin}\limits_{\omega} \sum_{i=1}^{|\mathcal{T}_{\textrm{train}}|} 
    \mathcal{L}\left(\theta_i^*(\omega), \mathcal{D}_i^{\textrm{val}}\right) }^{\text{outer level}}, \label{eq:outer-level}\\
    \text{s.t. } & \underbrace{\theta_i^*(\omega) = \argmin_\theta \mathcal{L}_i\left(\mathcal{D}^{\textrm{train}}_i, \theta,\omega\right)}_{\text{inner level}} \label{eq:inner-level}.
\end{align}

The inner level consists in learning task-specific learners conditioned on the meta-knowledge $\omega$. Note that the inner level only optimizes the task-specific parameters $\theta$ using the task train datasets $\mathcal{D}^{\textrm{train}}_i$ and does not change $\omega$. Whilst, the outer level learns $\omega$ that minimizes the aggregated losses from all the train tasks on their validation datasets. 

In the literature, it is common to divide meta-learning methods into three families: optimization-based, model-based, and metric-based. Optimization-based methods, promoted by the Model Agnostic Meta-Learning (MAML) algorithm \cite{DBLP:conf/icml/FinnAL17} have been recently adopted for domain generalization. The general idea is to consider the different domains as different tasks. Hence, data from multiple source domains are divided into meta-training and meta-testing sets. By training with data from different domains, the meta-learner is exposed to domain shift and is required to learn a meta-knowledge that quickly adapts to domain shift in new unseen domains \cite{DBLP:conf/aaai/LiYSH18}.

\subsection{Self-Supervised Learning}
Self-supervised learning (SSL) is a learning paradigm that generates labels from data and subsequently uses these labels as ground truth. SSL is useful in real-world applications where abundant unlabeled data is available, especially when the labeling process is cumbersome and expensive. Another motivation behind SSL is to learn rich and general representations, unlike supervised learning methods that learn biased representations via the supervision signal or the type of annotations \cite{DBLP:conf/cvpr/LiV19a}. In supervised learning, labels serve as the supervision signal to learn a specific task. However, in SSL, a model is learned using the data as a supervision signal. In other words, the labels in SSL are generated from the data itself. The SSL pipeline can be divided into two parts: 
\begin{itemize}
    \item learn feature representations by solving a \emph{pretext} task. An example of a pretext task is to retain part of the input data to be predicted by a model that is trained on the other part of the data \cite{MaskedAutoencoders2021}. Another pretext task consists in learning the relationship between data instances (e.g., similarity) or reconstruct an input from its shuffled parts (also known as the jigsaw puzzle). Note that the labels (or supervision signal) for the pretext task is generated from the input data, thus no human intervention is needed;
    \item solve a downstream task using the learned representations and a few annotated data.
\end{itemize}

SSL is applied in DG to learn domain-invariant features that help in avoiding overfitting on domain-specific biases while aligning features from different source domains. As discussed in Section \ref{subsec:data-invariant-representation-learning}, these invariant features can be leveraged in unseen target domains to achieve better generalization \cite{DBLP:conf/icml/MuandetBS13}. In this context, contrastive learning is a well-known SSL method that aims to learn latent representations such that positive instances are close and negative samples are pushed away. Therefore, in the learned embedding space, the distance between similar instances is reduced while the distance between negative pairs is increased. For instance, the authors in \cite{selfReg2021} proposed two self-supervised contrastive losses to measure feature dissimilarities in the embedding space. For dissimilarities across domains, the authors used a Mix-up layer \cite{mixup, DBLP:conf/icassp/WangLK20} (i.e., a convex combination of samples' embeddings from different domains) to compute the interpolated feature representation across domains. Thus, the regularisation loss is defined as the distance between the individual representations and the interpolated one using the mix-up layer. One caveat of this method is that it assumes the label space does not change for all the domains.

\subsection{Ensemble Learning}
Ensemble learning \cite{Zhou2012EnsembleMF} is a famous technique in traditional and modern machine learning where multiple models are learned and combined for prediction/classification. The same idea was also exploited for DG. The most straightforward approach is to learn a model for each source domain and average the individual predictions to compute the final ensemble prediction \cite{DBLP:journals/tip/ZhouYQX21, DBLP:conf/dagm/DInnocenteC18}. Instead of learning separate models for each source domain, it is common to design the ensemble as a shared feature extractor and different domain-specific heads \cite{DBLP:journals/tip/ZhouYQX21}. Another line of work focuses on the weighting of the individual models' predictions. For instance, the domain-specific models can be weighted differently depending on the similarity of the target domain to the source domain. The authors in \cite{DBLP:conf/icip/ManciniBC018} proposed to learn a domain predictor that predicts the probability that a target sample belongs to a source domain. These probabilities can be used to fuse the models' predictions at test time.

An alternative solution proposes to train domain-invariant classifiers for each source domain by learning domain-specific normalization \cite{DBLP:conf/eccv/SeoSKKHH20, DBLP:journals/pr/SeguTT23}. All the classifiers share the same parameters except the ones in the normalization layers. The objective of learning domain-specific normalization is to obtain domain-agnostic latent feature space that can be used to map samples from unknown domains to the source domains. This idea is related to the feature alignment methods reviewed in Section \ref{subsubsec:explicit-feature-alignment}.

Alternatively, the stochastic weight averaging (SWA) method \cite{SWA} aggregates weights at different training epochs to form an ensemble model instead of combining the predictions of multiple learners. Starting from a pre-trained model, SWA trains a single model using a cyclic learning rate schedule (or a constant high learning rate) and saves model snapshots corresponding to different local minima. Averaging these points leads to better solutions in flatter regions of the loss landscape. Intuitively, flatter minima are more robust than sharp minima to changes in the loss landscape between the training and testing datasets \cite{SWA}. Consequently, this weight averaging idea was extended to the DG proving that flat minima lead to better generalization on unseen domains \cite{SWAD}. 

\subsection{Hypernetwork-Based Learning}
Hypertnetwork-based learning \cite{hypernetworks} is an approach that learns a network (i.e. the hypernetwork) to generate weights for another network called the main network. The latter represents the usual model that maps raw data to their targets or labels. The goal of the hypernetwork is to generate a specific set of weights depending on inputs about the structure of the weights or tasks. Different from the usual supervised learning setting, only the hypernetwork's parameters are learned during training whilst keeping the main network's parameters unchanged. At inference, the main network is evaluated based on the weights generated by the hypernetwork.  

Recent work proposed hypernetwork-based algorithms for DG in natural language processing \cite{DBLP:journals/corr/abs-2203-14276} and vision \cite{hmoe}. For vanilla DG, a straightforward application of hypernetworks is to train a hypernetwork on data samples from different source domains to produce the model's weights for each domain. On the other hand, for compound DG, the appropriate approach is to first learn a latent embedding space for the different domains, then the hypernetwork learns to map the latent features to a set of weights so as to compute model predictions. 

In the next sections, we will overview the different applications of the techniques detailed above to wireless communication problems.  

\section{Domain Generalization Applications in Wireless Communications}\label{sec:future-direction-applications}

When designing data-driven ML-based algorithms for solving wireless communication problems, it is crucial to ensure that the developed algorithms have guaranteed generalization capabilities. However, little effort has been devoted to investigating the DG issue despite the huge research effort in applying data-driven machine learning techniques to various wireless communication. The goal of this section is to overview the existing DG methodologies that were applied by the communication community, and summarize the learned lessons from their applications.

\subsection{Channel Decoding}

Iterative turbo/LDPC decoders \cite{DBLP:journals/tcom/BerrouG96,DBLP:journals/tit/RichardsonSU01} based on the belief-propagation (BP) framework \cite{pearl1988probabilistic} are recognized as state-of-the-art channel decoders because of their capacity approaching/achieving performance for relatively large block lengths. For this reason, they have been adopted in the 4G/5G communication standards.

Many deep learning studies have shown that data-driven ML techniques can decrease the BP decoding complexity especially for short-to-moderate block lengths \cite{DBLP:conf/iccchina/NiuDTG21}. For short-block-length polar codes \cite{DBLP:journals/tcom/Trifonov12} (e.g., 16 bits), DNN-based decoders were shown to exhibit near-optimal performance using maximum a posteriori (MAP) decoding \cite{DBLP:conf/ciss/GruberCHB17}. For larger block length codes (i.e., larger than 100 bits), the BP algorithm was unfolded into a DNN in which weights are assigned to each variable edge, thereby showing an improvement in comparison to the baseline BP method \cite{DBLP:conf/allerton/NachmaniBB16}. By varying the signal-to-noise ratio (SNR) values of the received signal, hypernetworks have been employed to generate the weight of a variable-node network in the Tanner graph \cite{DBLP:conf/nips/NachmaniW19}. Together, all the variable-node networks represent the graph neural network (GNN) on which message passing is performed. Meta-leaning algorithms have been explored in \cite{DBLP:conf/nips/LiBMK0LH21} as part of an end-to-end learning approach. There, meta-tasks were designed by varying the SNR to account for the task difficulty, under a convolutional encoder with a fixed coding rate of $1/2$.

Overall, the aforementioned ML-based channel decoding methods can be classified into two categories \cite{DBLP:conf/iccchina/NiuDTG21}:
\begin{itemize}
    \item \textit{Data-driven methods}: these methods promote end-to-end learning approaches by substituting all the BP decoding components with a DNN \cite{DBLP:conf/ciss/GruberCHB17}. Here, the structure of the code is ignored, and the channel decoding problem is regarded as a classification task from the input (i.e., received signal) to the output (i.e., decoded bits).    
    \item \textit{Model-driven methods}: the goal of this family of methods is to substitute the decoding components of the classical BP-based decoder (e.g., deinterleaver, log-likelihood ratio estimators) with trained DNNs without altering the classical sequence of decoding components \cite{DBLP:conf/nips/JiangKAKOV19,DBLP:journals/jstsp/NachmaniMLGBB18,DBLP:conf/ita/VasicXL18}.
\end{itemize}
Little attention has been, however, paid to studying how DG methodologies can be applied to both categories beyond the simple variation of the SNR values. Moreover, empirical and theoretical understanding of their potential for channel decoding is still lacking.

\subsection{Channel Estimation}

One of the crucial components of any wireless communication system is the channel estimator \cite{heath2018foundations}. A vast body of prior work made use of data-driven ML techniques for channel estimation to show the attractive features of DNNs such as the low computational complexity at inference time \cite{DBLP:journals/wcl/YeLJ18,DBLP:journals/twc/HuGZJL21,DBLP:journals/icl/SoltaniPMS19,gizzini2020deep}. None of these studies, however, did analyze the impact of the distribution shifts on the reported estimation performance. Indeed, little effort has been devoted to investigating the robustness of DG algorithms in estimating wireless channels.

Another channel estimation algorithm for wideband mmWave systems was proposed in \cite{CE-ISTA-hypernetworks} based on unfolding the iterative shrinkage thresholding algorithm with a few learnable parameters. This algorithm was further extended to include a hypernetwork for the sake of generalization to new environments. Given the SNR level and the number of resolvable paths, the hypernetwork generates suitable learnable parameters for the channel estimation model. Alternatively, in \cite{HypetNetCE}, the authors proposed to train a hypertnetwork to learn weighting factors so as to aggregate channel estimation models learned for three main scenarios: urban micro, urban macro, and suburban macro. Hypernetwork recurrent DNNs have also been used to track wireless channels over a wide range of Doppler values \cite{DBLP:conf/globecom/PratikABSW21}. For this multi-Doppler case, classical tracking methods make use of a bank of Kalman filters with an additional Doppler estimation step. Meta-learning was also adopted to train an encoder-decoder architecture to quickly adapt to new channel conditions by varying the number of pilot blocks preceding the payload in each transmission block \cite{DBLP:conf/spawc/ParkSK20}. For sparse MIMO channel estimation, the optimization/estimation modules of the approximate message passing (AMP) \cite{donoho2011design} and vector AMP (VAMP) \cite{DBLP:journals/tit/RanganSF19} algorithms were substituted by learnable DNNs \cite{DBLP:journals/tsp/BorgerdingSR17}. Specifically, DNNs did not neglect the ``Onsager correction'', which lies at the heart of the AMP paradigm, and was rather employed to construct the underlying DNNs. By doing so, it was shown that the Onsager correction is beneficial to train DNNs that $i)$ require fewer layers to reach a predefined level of accuracy and $ii)$ yield greater accuracy overall as compared to DNNs ignoring the Onsager correction term.

Designing multiple channel estimation tasks pertaining to distinct domains requires varying wireless transmission parameters to simulate different channel communication scenarios. As depicted in Fig.~\ref{fig:domain-randomization}, these parameters are categorized as:

\begin{itemize}
    \item \textit{Propagation parameters} which capture the different types of randomness in channel models \cite{heath2018foundations}. They are not under control in practical communication scenarios.
    \item \textit{System parameters} which govern multiple aspects of communication systems that are set by system designers such as the code rate, the number of transmit and receive antennas, the type and order of the modulation constellation, and the carrier frequency, etc.
\end{itemize}
\noindent It is worth noting that varying these parameters to generate different domains will lead to one or multiple types of distribution shifts. As one example, the design of a channel estimator for broadband communication has to generalize over the channel distributions. With the widely adopted strategy for bandwidth expansion, known as carrier aggregation \cite{DBLP:journals/cm/YuanZWY10}, the distribution of the channel coefficient shifts across multiple non-contiguous narrow frequency bands. For this reason, assuming that the channel is the output of a DNN, the DNN-based channel estimator has to account for the label shift of the estimated channel coefficients because their support changes as a function of the frequency band.

Other related studies focusing on continual learning (CL) benchmarked the performance of CL-based methods for MIMO channels estimation by varying the SNR and the coherence time of the channel \cite{akrout2022continual}. A continual learning minimum mean-square error (CL-MMSE) method has also been proposed in \cite{kumar2021continual} where the DNN adapts to different numbers of receive antennas between 8 and 128 to generate tasks with different difficulties.

\subsection{Beamforming}

Steering the main lobe of antenna array systems toward users in a real-time manner (i.e., beamforming) is a critical task to minimize interference and enhance the achievable rate of wireless communication systems. This is because, the antenna array processing in adaptive/reconfigurable digital signal processing algorithms assume no mismatch between the actual and expected array responses to the received signal \cite{monzingo2004introduction}. With the increase of the number of antenna elements in massive MIMO systems, a larger number of degrees of freedom is achieved at the cost of higher algorithmic complexity incurred when optimizing the beamformer weights \cite{heath2018foundations}. Since beamforming weights must be continuously computed under changing propagation environments, ML methods have been explored as a possible solution to low-complexity beamforming design \cite{DBLP:conf/temu/ZaharisYSXLMM16,zaharis2020effective}. For instance, the weighted minimum mean-square error (WMMSE) estimator of the transmit MISO beamforming vector was unfolded such that each estimation iteration corresponds to a DNN \cite{DBLP:journals/ojcs/PellacoBJ22}. By doing so, the matrix-inverse operation of the standard WMMSE estimator is avoided in addition to the advantage of a lower computational complexity without sacrificing the estimation performance. It was also reported that fully distributed reinforcement learning (RL) estimates the uplink beamforming matrix by dividing the beamforming computations among distributed access points without significant accuracy deterioration \cite{DBLP:journals/tccn/FredjAMAH22}. We refer the reader to \cite{DBLP:journals/access/KassirZLKYX22} for a comprehensive review of ML-based beamforming methods.

Few studies, however, have considered DG as an important ingredient to assess the performance of ML-aided beamforming solutions based on the meta-learning framework reviewed in Section \ref{subsec:meta-learning}. A meta-learning algorithm for weighted sum rate maximization was proposed for beamforming optimization in MISO downlink channels \cite{DBLP:conf/isit/XiaG21}. Instead of using the WMMSE algorithm iteratively to update each variable involved in the beamforming optimization problem, long-short-term-memory (LSTM) networks were used in the inner-loop of the meta-learning framework to learn the dynamic optimization strategy and hence update the optimization variables iteratively. The outer-loop of the meta-learning framework, however, makes use of the updated parameters to maximize the weighted sum rate. This strategy adaptively optimizes each variable with respect to the geometry of the sum-rate objective function, thereby achieving a better performance than the WMMSE algorithm. Another line of work employed the standard meta-learning MAML algorithm \cite{DBLP:conf/icml/FinnAL17}
for adaptive beamforming to new wireless environments \cite{DBLP:journals/twc/YuanZWOL21}. This work was further extended to reduce the complexity of the MAML algorithm by dedicating a DNN model as a transferable feature extractor for feature reuse across wireless channel realizations \cite{DBLP:journals/twc/ZhangYZKW22}. Self-supervised learning was used to map uplink sub-6 GHz channels into mmWave beamforming vectors without accessing labeled training datasets \cite{DBLP:journals/twc/ChafaaNBD22}. By exploiting a dataset containing pairs of uplink and downlink channels, DNNs learned implicitly and autonomously the data representations from correlations in the training data pairs to predict the beamforming vectors.

\subsection{Data Detection and Classification}

To decrease the computational complexity of classical data detection algorithms, ML techniques were proposed to detect communication signals under various conditions by reformulating bit/symbol detection as a conventional classification problem \cite{DBLP:journals/jstsp/DornerCHB18,DBLP:journals/corr/FarsadG17,DBLP:conf/spawc/SamuelDW17,al2019learning}. In this context, the various DG techniques reviewed in Sections \ref{sec:data-manipulation}--\ref{sec:learning-paradigms} can be leveraged to investigate the generalization capabilities of DNNs when applied to the data detection problem. For instance, DG demodulation methods for \textit{multiple} modulation schemes have to account for both concept and label shifts of the estimated symbols because the modulation constellation varies from one domain to another. This scenario corresponds to wireless transmissions with adaptive modulation and coding where the choice of modulation order and coding rate is based on the instantaneous channel quality indicator (CQI).

Recently, data detection in MIMO systems with spatially correlated channels has been extensively studied. Indeed, MMNet \cite{mmnet} proposed an unfolding algorithm based on approximate message passing augmented by learnable parameters to achieve state-of-the-art performance on correlated channels. However, this algorithm needs to be re-trained for each channel realization. To overcome this drawback, the authors proposed to use a hypernetwork to predict the learnable parameters based on perfect CSI and noise power knowledge \cite{HypernetSD}. The generalization of this framework was tested under different SNR levels and user mobility settings to simulate different channel spatial correlations. One drawback of this approach is that it assumes that the CSI and noise power are perfectly known at the receiver. Similarly, the unfolded version of the expectation propagation detector was proposed wherein damping factors are learned using meta-learning \cite{zhang2020metaEpNet}. This detector was also extended using hypernetworks to achieve generalization to new channel realizations and noise levels but for typical values of many other system parameters \cite{zhang2021EPNetHypernetwork}. The major drawback here is that DNN must be retrained for each set of new system parameters. A meta-learning strategy was also used to train the damping factors of the VAMP algorithm to improve its convergence speed and quickly adapt to new environments, thereby yielding more accurate signal detection performance \cite{DBLP:journals/twc/ZhangHLWJ21}.

Other similar types of detection/recognition tasks are also of the same classification nature such as modulation classification in non-cooperative communication systems \cite{o2016convolutional} and wireless transmitter classification \cite{youssef2018machine}. These works focus on improving the classification accuracy only, and the generalization ability of DNNs was studied in a few prior work only \cite{DBLP:journals/twc/LiuOP22}.

\subsection{Beam Prediction}

Since 6G and beyond communication systems are moving to higher frequency bands (e.g., mmWave and sub-terahertz), developing techniques for narrow directive beam management is critical to guarantee sufficient receive power. Existing solutions rely on leveraging the channel sparsity \cite{DBLP:journals/jstsp/HeathPRRS16a}, constructing adaptive beam codebooks \cite{DBLP:journals/tcom/AlrabeiahZA22}, and beam tracking \cite{DBLP:journals/icl/JayaprakasamMCK17}. Due to beam training overheads, these classical strategies, however, cannot meet the ever-increasing data rate demands of emerging applications for future systems with large antenna arrays serving highly-mobile users and latency-critical devices \cite{DBLP:journals/corr/abs-2111-11177}. For these reasons, the development of ML-aided methods can offer data-driven solutions for the beam management problem because the beam direction decision depends on the user location and the geometry of the surroundings about which sensory datasets can be collected.

A practical ML solution is expected to generalize to unseen scenarios and operate in realistic dense deployments. The fact that practical sensors do not normally provide accurate enough positions/orientations for narrow beam alignment motivates acquiring multi-modality datasets about the environment such as sub-6GHz channel information, LiDAR point clouds, and radar measurements \cite{DBLP:journals/corr/abs-2209-07519}. DG algorithms should be developed to leverage these datasets representing different domains in the same environment. For example, ideas from domain-invariant representations are beneficial to cope with distribution shift sources such as the quality of collected measurements (e.g, noise level, sensitivity to weather conditions), user mobility, and signal blockages. These factors lead to the acquisition of multiple data domains which can be exploited to learn both domain-invariant and domain-specific features to determine the index of the optimal beamforming vector from the codebook in a generalizable manner.

\subsection{RIS-Aided Wireless Communications}

Wireless communications aided by RISs has triggered a remarkable research effort in the last few years \cite{DBLP:journals/tsp/HuRE18a}. The possibility to purposely manipulate the electromagnetic propagation environment via the use of IRSs, incorporation of IRSs as integral part has pushed researchers to revisit fundamental wireless communication problems (e.g., beamforming, channel estimation) and incorporate the impact of RISs on the overall communication system performance measured in terms capacity, estimation accuracy, secrecy, outage, and energy efficiency. In this context, ML methods belonging to multiple learning paradigms (e.g., supervised/unsupervised learning, reinforcement learning, federated learning) have been also devised to account for the propagation effects of IRSs. We refer the reader to the survey in \cite{DBLP:journals/access/FaisalC22} for an exhaustive summary of ML approaches for RIS-aided communication.

In regard to DG, only a handful of studies have assessed the performance of ML methods from the perspective of accuracy-generalization tradeoff. The problem of channel estimation for RIS-aided communication has been investigated in \cite{DBLP:journals/icl/TsaiCTW22} where an adaptive shrinkage parameter
based on a hypernetwork was used instead of a fixed shrinkage parameter. Based on the current channel recovery status, the hypernetwork provides an updated shrinkage parameter thanks to which the IRS-aided channel estimation accuracy has been assessed over different iterations as well as SNR values ranging between -10 dB and 25 dB. This work does not study DG as a function of the wireless communication parameters but rather with respect to the algorithmic steps of the LAMP algorithm. The robustness to additional noise of RL algorithms when the CSI is perturbed has been examined in \cite{DBLP:journals/corr/abs-2107-08293} in the context of the optimization of RIS phase shifts. This work showed that RL methods exhibit resilience to different channel impairments as compared to classical optimization methods in the evaluation step only. In other words, DG training methodologies were not adopted and hence the work does not consider handling the domain shifts in estimating the phase shifts and only reports the performance degradation during inference.

\subsection{Applications in Edge Networks}

The domain shift problem arises naturally in IoT applications due to the heterogeneity in devices' behavior,  spatial and temporal information, etc. For healthcare IoT sensors, the work in \cite{iotalignmentcovid} applied a data alignment algorithm to learn and project accelerometer data from different users into a common feature space. The learned shared feature space is then used to track users' symptoms. 
For vehicule-to-everything (V2X) applications, a meta-learning approach for power allocation tasks has been proposed in \cite{DBLP:journals/tvt/YuanZWL21} to enhance DNNs to achieve fast adaption to new environments with limited interactions.

DL has been applied in human activity recognition to extract meaningful features from raw sensory data instead of hand-engineered ones. Human activity recognition usually involves multi-modal sensory data from multiple devices/subjects to predict one or multiple activity labels. For the same activity, sensor data can vary depending on the subjects' characteristics such as gender, age, and behavior. One solution to this intra-activity shift problem is to remove the user-specific feature from the sensory information and keep the common activity features across all users only. To do so, feature disentanglement is proposed to learn two groups of representations: the common activity features and user-specific representations \cite{su2022learning}. Another line of work focused on learning statistical features from sensory data using kernel-based techniques \cite{kernelHAR2, kernelHAR}. These studies, however, make use of kernel-based methods for more predictive feature extraction from raw sensory data only. The use of kernel-based methods to improve DG as explained in Section \ref{sec:representation-learning} was not explored.

 \subsection{Summary and Lessons Learned}
In the preceding sections, we reviewed different key applications in wireless communication where DG algorithms should be further investigated for the sake of robust generalization. Our observations and lessons learned are summarized below.
\begin{itemize}
    \item \textbf{Lack of DG algorithms for wireless}: To better judge the suitability of data-driven ML methods for real-world communication uses cases, it is crucial to determine the uncertainty of ML algorithms and analyze their ability to generalize in order to lay the ground for rigorous evaluation protocols. However, minimal effort has been dedicated by the communication community to initiate such an investigation. As one example, use cases in 3GPP Release 18 package such as CSI compression with autoencoders raise multiple interesting DG questions. Questions about the autoencoder training procedure, as well as, the different user traffic scenarios and urban areas should be considered before determining the source and target datasets.
    \item \textbf{One-sided focus on end-to-end DG}: Most DG algorithms are produced by the ML community and hence lack wireless communication knowledge in their designs. As a consequence, most DG communication papers make use of ML end-to-end techniques that are blind to the characteristics of the communication problem at hand. While this trend is worthwhile to assess the generalization performance of end-to-end learning methods, tailoring existing DG algorithms and devising new ones are essential research avenues that require further investigation.
    
    \item \textbf{Need for wireless DG benchmarks}: The number of DG benchmarks within the ML community has significantly increased over the last few years due to the need for algorithmic generalization evaluation (see \cite{DBLP:conf/ijcai/0001LLOQ21} for a comprehensive review). Unfortunately, with few exceptions \cite{DBLP:journals/corr/abs-2209-07519,DBLP:conf/nips/LiBMK0LH21}, the absence of a unified benchmarking in wireless communications renders the comparison of the different proposed DG algorithms impossible. Consequently, it is crucial to establish a unified framework to analyze the improvements of the research endeavors and henceforth design robust and efficient DG algorithms.    
\end{itemize}

\section{Open Issues}\label{sec:future-direction} 
In this section, we discuss some of the open questions that evolve around the necessity to carefully incorporate the DG concept in ML-oriented communication research. This is because, unlike many other ML-based technologies, most real-world communication applications require real-time operation and seamless adaptation to dynamically changing propagation conditions. This precludes the luxury of repeatedly training ML-oriented models and makes DG-induced robustness a must-have feature in any ML-aided communication system.

\subsection{Beyond End-to-End Learning for Generalization}

Most of the existing studies rely on end-to-end learning to train a holistic over-parametrized DNN architecture by applying gradient-based optimization to the learning system as a whole. This means that all transmit/receive modules of the communication system must be differentiable (in the reverse-mode algorithmic differentiation sense \cite{speelpenning1980compiling}). Few wireless communication libraries have been proposed to study differentiable communication systems \cite{sionna, AIwirelessMatlab}.

\begin{figure*}[!b]
\centering
\begin{minipage}[c]{\textwidth}
\centering
\includegraphics[scale=0.71]{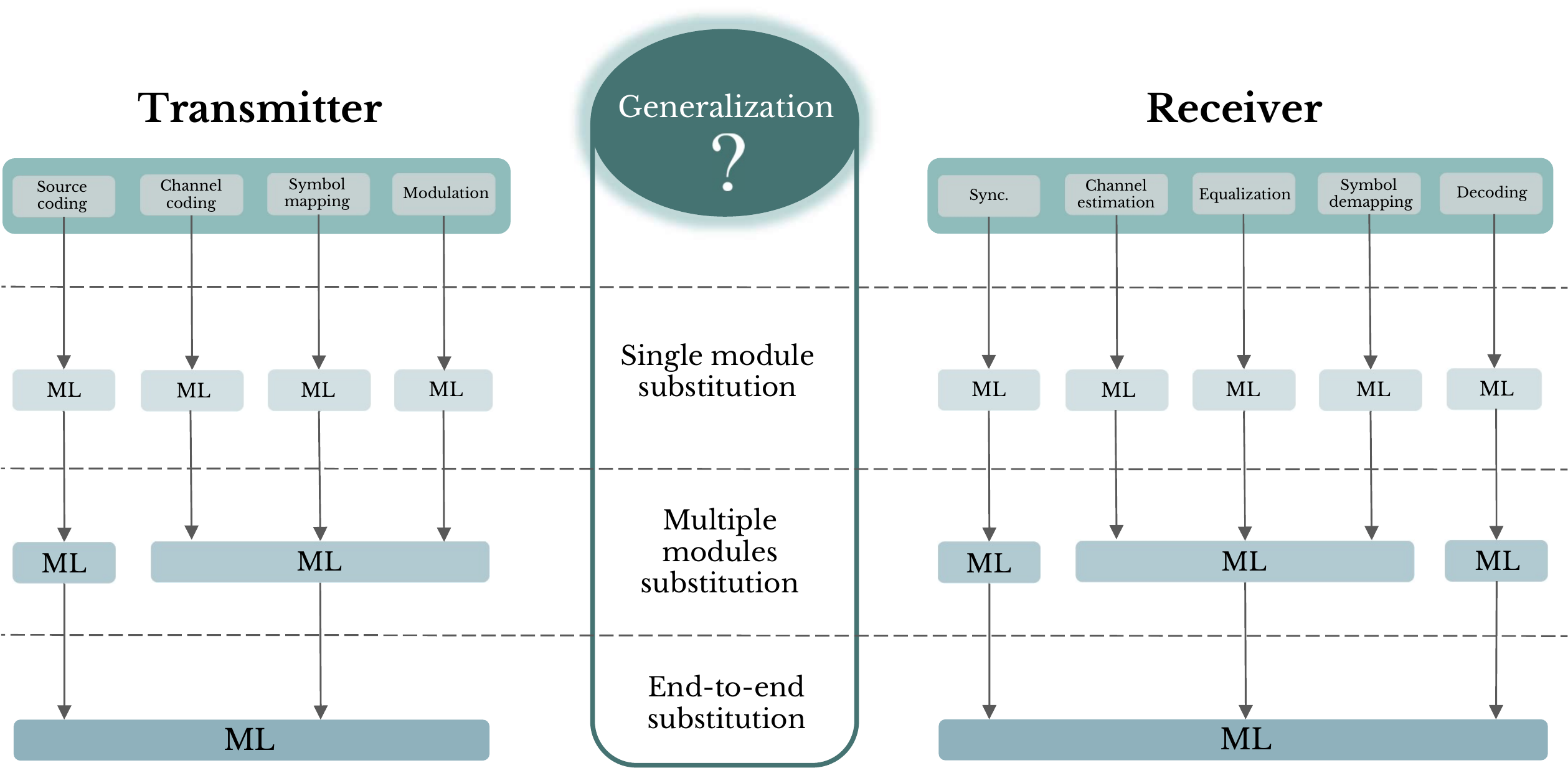}
\caption{The possible integration steps of ML methods into the conventional transmit/receive communication chain if ML methods will be proven to be robust to domain shifts.}
\label{fig:end-to-end-AI-receiver}
\end{minipage}
\end{figure*}

Before advocating the adequacy of applying ML methods to the building blocks of the wireless physical layer depicted in Fig.~\ref{fig:end-to-end-AI-receiver}, DG has to be meticulously investigated and guaranteed \textit{within} and \textit{across} the blocks. From this perspective, it is not enough to claim the migration from model-based classical signal processing techniques to data-driven ML techniques without analyzing the impact of each migration on the overall system performance in terms of both accuracy and robustness. While such migration is a conceptually profound paradigm shift, its impact continues to be assessed from the accuracy perspective only, and hence must also be carefully analyzed through the lens of generalization/robustness.

The legacy physical layer design strategy relies on the divide-and-conquer approach by decomposing (a.k.a. layering) the entire communication chain into smaller blocks \cite{gallager2008principles}. Designing ML methods to substitute a single block or multiple blocks (see Fig.~\ref{fig:end-to-end-AI-receiver}) raises critical generalization questions justified by the following two facts:
\begin{itemize}
    \item End-to-end learning methods are trained with gradient descent-like optimizers, which exhibit slow convergence on ill-conditioned problems or convergence to possibly poor local optima. In other words, training is performed while hoping that the structural preconditioning is sufficiently strong to steer a method as simple as gradient descent from a random initial state to a highly non-trivial solution \cite{DBLP:conf/acml/Glasmachers17}. This assumption is risky since all ML techniques tailored for wireless applications are exclusively used for non-convex optimization problems.
    \item The valuable wireless communication know-how developed since the 50s is completely neglected during end-to-end training. ``Standing on the shoulders of giants'' (as Sir Isaac Newton once said) is a scientific tradition which promotes building upon the accumulated knowledge and discoveries made by others, and ``end-to-end learning'' must be proven robust to domain shifts to be considered an exception.
\end{itemize}

For these considerations, going beyond conventional end-to-end learning is an important step towards answering critical DG questions in data-driven ML techniques applied to wireless communications. In what follows, we discuss research
directions to cope with some end-to-end learning limitations.

\subsection{Hybrid Data-Driven and Model-Driven Methods}

After more than a century-long research effort in radio communications, state-of-the-art communication modeling and fast estimation algorithms are becoming more essential to high-bandwidth transmissions. From a DG perspective, the power of these classical model-driven tools lies in their guaranteed generalization capabilities because they do not depend on specific domains that are tied to generated/collected datasets. This generalization, however, often comes at the cost of high complexity.

Data-driven methods can come into play as an effective tool to reduce the computational complexity of classical model-based methods at the cost of generalization. As advocated in \cite{pellaco2022machine}, a hybrid framework that combines the benefits of both data-driven and model-based techniques is worth pursuing. Adopting this framework will prevent the generated domains for DG from being fully dependent on $i)$ the convergence of gradient-based optimizers for data-driven methods, or $ii)$ the complexity of model-based methods. For better illustration, we elaborate in what follows on how data-driven methods can be combined with physically consistent model-based methods.

The study of DG for MIMO communication should benefit from the side information provided by the physical laws governing the wave transmission and the circuits of RF components (i.e., amplifiers, and antennas). By employing physically consistent models \cite{DBLP:journals/tcas/IvrlacN10,DBLP:journals/corr/abs-2208-01556,DBLP:journals/twc/PizzoSM22}, it is possible to exploit the inherent symmetries and invariances in communication scenarios owing to Maxwell's equations \cite{jackson1999classical,baum1995symmetry}. From this perspective, physically consistent models for wireless communications offer an opportunity to generate communication datasets which exhibit domain-invariant regularities (e.g., antenna impedances), thereby diminishing the generalization difficulties across domains. As one example, fixing the impedance matrices of transmit and receive linear/planar antenna arrays increases the amount of correlation in the wireless channel, which can be exploited by DNNs for better channel estimation accuracy.

Moreover, this physically consistent direction opens the door for the analysis of DG through the lens of antenna theory. For example, it might be possible to determine which spacing parameter of the antenna array provides the best DNN accuracy for channel estimation. By doing so, realistic wireless communication domains are generated and more faithful representations of the real-world transmissions are simulated, thereby leading to a physically consistent version of digital twins for wireless communications \cite{DBLP:journals/cm/KhanSNHH22}.

\subsection{From Image-Based DG Methods to Signal-Based Methods}

Existing DG methodologies have been predominantly geared towards image-based vision tasks, leaving signal-based tasks almost unexplored despite being versatile in several real-world applications such as healthcare, retail, climate, finance, and communication. This unbalanced exploration impacts the development of specific DG methods for signal-based tasks. For instance, feature alignment approaches for DG are relying heavily on DNNs as feature extractors which are specifically fine-tuned to vision tasks, thereby leaving DG feature extraction for non-image signals severely underexplored. Some work looked at temporal distributional shifts in clinical healthcare \cite{guo2022evaluation,DBLP:conf/iconip/MaLZL19} and climate \cite{DBLP:conf/nips/MalininBGGGCNPP21} applications, but none of the prior work explored it in wireless communication. 

From this perspective, we highlight the importance of taking the first step towards a deeper understanding of temporal distributional shifts in wireless communication due to dynamic changes in the received signal resulting from the varying propagation properties (e.g., coherence time and Doppler shift).

\subsection{Compound Domain Generalization}
As mentioned previously, most of the presented methods for DG assume a homogeneous setting where domain labels are available. However, this assumption may not be realistic in several problems where the domain labels are hard to obtain or define. In this case, several techniques discussed above either become inapplicable (e.g., meta-learning) or their performance degrades drastically \cite{DBLP:journals/corr/abs-2103-02503}. Recently, there has been a surge of interest in studying the compound DG setting in vision problems. Most of the methods for compound DG propose to infer latent domain information from data and then use standard learning techniques to generalize across the latent domains. These solutions are, however, based on different restrictive assumptions such as: $i)$ the latent domains are distinct and separable \cite{hmoe}, $ii)$ the domain heterogeneity originates from stylistic differences \cite{DBLP:conf/cvpr/ChenL0LY22} or $iii)$ the latent domains are balanced \cite{DBLP:journals/corr/abs-1911-07661}. Compound DG is hence still an active research field with a lot of room for improvement, especially in wireless communication problems.

\subsection{Federated Domain Generalization}

Distributed learning algorithms enable devices to cooperatively build a unified learning model across agents with local training. As a result, a wide variety of distributed ML methods have been proposed and extensively analyzed within the federated learning (FL) framework \cite{DBLP:journals/corr/KonecnyMYRSB16}. 

For wireless physical layer applications, FL has been explored to address multiple key communication problems beyond the data security aspect \cite{DBLP:conf/infocom/ZhangYD22} such as channel estimation \cite{DBLP:journals/twc/ElbirC22}, symbol detection \cite{DBLP:conf/ssp/MashhadiSEG21} and beamforming \cite{elbir2020federated}. All of these works do not assume the availability of a central entity (e.g., base station) at which the learning model is trained. However, the question of whether the model learned by each agent generalizes to unseen scenarios is still unanswered and this remains an unexplored research area. In the context of IoT applications, very few efforts started investigating the challenges of DG for IoT devices by aligning each device's domain to a reference distribution in a distributed manner \cite{zhang2023federated}.

Addressing DG in the FL context is known as \textit{federated domain generalization} (FDG)\cite{DBLP:journals/corr/abs-2111-10487}. Distributed agents can collect their local data independently, hence naturally forming a distinct source domain. At the time of writing, no research paper in wireless communication has studied FDG, e.g., in the context of distributed MIMO \cite{DBLP:journals/jsac/WangWYWCH13} consisting of distributed antenna array systems.

\section{Conclusion}\label{sec:conclusion}

Studying the impact of distribution shifts on the performance of ML-based algorithms for wireless applications is of paramount importance to our research community to better reflect on the adequacy of adopting the data-driven ML approaches in communication systems engineering. In particular, the investigation of domain generalization will lay the ground for rigorous evaluation protocols of data-driven algorithms for wireless communications systems. In this paper, we presented an overview of state-of-the-art methodologies for domain generalization problems to handle distribution shifts. To justify the need to devise new algorithms with better generalization capabilities, we distinguished the four types of distribution shifts between source and target domains. We also provided an overview of  multiple important fields related to generalization to better put domain generalization in proper perspective across close research areas. Then, we summarized the three existing methodologies to improve the generalization capabilities of deep learning models, namely, data manipulation, data representation, and domain generalization learning paradigms. In doing so, we gave multiple examples and suggestions not covered in the current literature where these methodologies can be applied for wireless communications applications. We then reviewed the recent research contributions to improve the generalization of neural network models when solving wireless communication problems. These problems involve beam prediction, data detection, channel decoding, beamforming, edge networks, etc. We also presented the learned lessons from the existing applications of domain generalization methodologies for wireless communication problems by highlighting the lack of $i)$ algorithms exploiting the domain knowledge from well-established communication models, and $ii)$ open-source benchmarks to accelerate the development of robust algorithms for future wireless networks. Finally, we discussed open questions to enrich and bridge the gap between both domain generalization and wireless communication fields.  

\bibliographystyle{ieeetr}
\bibliography{refs}

\end{document}